%% file: acl.tex
\pdfoutput=1

\documentclass[11pt]{article}

\usepackage[]{acl}

\usepackage{times}
\usepackage{latexsym}
\usepackage{graphicx}
\newcommand{\change}[1]{{\leavevmode\color{black}#1}}
\usepackage[T1]{fontenc}

\usepackage[utf8]{inputenc}

\usepackage{microtype}

\title{Domain Adaptation with Pre-trained Transformers for \\ Query Focused Abstractive Text Summarization}

\author{
Md Tahmid Rahman Laskar\textsuperscript{1,2,}\thanks{\hspace{0.115cm}Work being done when the author was at York University.} , Enamul Hoque\textsuperscript{3}, Jimmy Xiangji Huang\textsuperscript{2,3} \\
    \textsuperscript{1}Dialpad Canada Inc. \\
  \textsuperscript{2}Information Retrieval \& Knowledge Management Research Lab, York University \\
  \textsuperscript{3}School of Information Technology, York University \\
  Toronto, Ontario, Canada \\
 \texttt{tahmedge@cse.yorku.ca}, \texttt{enamulh@yorku.ca}, \texttt{jhuang@yorku.ca} \\}

\begin{document}
\maketitle
\begin{abstract}
The Query Focused Text Summarization (QFTS) task aims at building systems that generate the summary of the text document(s) based on the given query. A key challenge in addressing this task is the lack of large labeled data for training the summarization model. In this paper, we address this challenge by exploring a series of domain adaptation techniques. Given the recent success of pre-trained transformer models in a wide range of natural language processing tasks, we utilize such models to generate abstractive summaries for the QFTS task for both single-document and multi-document scenarios. For domain adaptation, we apply a variety of techniques using pre-trained transformer-based summarization models including transfer learning, weakly supervised learning, and distant supervision. Extensive experiments on \change{six} 
datasets show that our proposed approach is very effective in generating abstractive summaries for the QFTS task while setting a new state-of-the-art result in \change{several datasets} across a set of automatic and human evaluation metrics.

\end{abstract}
\input{tables}

\input{figures}

\section{Introduction}
\label{section:intro}

With the rapid growth of textual documents on the internet, accessing information from the web has become a challenging problem \cite{yao2017survey}. In a web search, users may require the summary about a certain topic from various sources to fulfill their information needs \cite{xulapata2020coarseemnlp}. Since the performance of the web search engines largely depends on a system that possesses good question answering (QA) capabilities, many researchers are focusing on developing  systems that can provide users with a summarized response to their queries \cite{deng2020jointanswerselectionsummarygeneration}. The \textit{Query Focused Text Summarization} task deals with such problems, where a query along with the source document(s) are given and the objective is to generate a summary from the source document(s) based on the given query \cite{yao2017survey} (see Table \ref{tab:introduction}). 

The query focused summarization task can be categorized depending on the type of the source document(s) and the generated summary. For instance, based on the type of the source document(s), we can consider two scenarios: (i) \textit{Single-Document Scenario:} where the goal is to generate a summary from a single source document, and (ii) \textit{Multi-Document Scenario:} where the goal is to generate a summary from a set of documents \cite{baumel2018query}. Moreover, based on the type of the generated summaries, this task can be either extractive or abstractive \cite{baumel2018query,nema,queryfocusedsummarization2017unsupervised,yao2017survey,xie2020conditionalmicrosoftqfas,pasunuru2021data}. For the \textit{Extractive Summarization Scenario}, relevant text spans are directly extracted from the source document(s). In contrast, for the \textit{Abstractive Summarization Scenario}, the generated summaries can contain words the may not appear in the source document(s). 

Given the rise of conversational QA assistants such as Siri, Cortana, Alexa, and Google Assistant, researchers are interested to study how to incorporate abstractive summarization capabilities in such systems for natural response generation \cite{nishida2019multi}. Due to the growing interest on QA systems with summarization capabilities, a number of methods have been proposed for the Query Focused Abstractive Summarization (QFAS) task. More recent methods for such tasks adopted various state-of-the-art neural summarization models \cite{yao2017survey,2020prenlpsurvey} by following the encoder-decoder architechture.

However, there are some key challenges that are required to be addressed while building QFAS systems for both single and multi-document scenarios. 
For the single-document QFAS task, one major challenge is that the available datasets are very small in size compared to the generic abstractive summarization datasets \cite{nema,baumel2018query,see}. Thus, during training, the model needs to tackle the few-shot learning problem. 

For the multi-document scenario, again the existing benchmark datasets are very small \cite{baumel2018query}. On top of that, each gold reference summary in the available datasets are written for a given document set without including the reference summary of each individual document in that document set. The problem is that we cannot simply concatenate all the documents in a given document set and feed into the state-of-the-art neural architecture for text summarization: the transformer model \cite{liuemnlpbertsum,zhang2019pegasus,bart,t5}, as the input sequence may become prohibitively long. This is because the transformer architecture has quadratic computational time and memory complexities and these complexities get worsen with the increasing length of the input sequence due to the matrix multiplication in self-attention blocks \cite{2020reformer,beltagy2020longformer,zaheer2020bigbird,wang2020linformer,choromanski2020rethinkingperformer}.
 
To address the above challenges, in this paper, we study how to utilize domain adaptation from pre-trained neural models for the QFAS task. Note that domain adaptation or transfer learning from pre-trained models is particularly suitable 
when the target dataset does not contain any labeled training data or the size of the training dataset is very small \cite{ramponi2020neuraldomainadaptationsurvey}. \change{To leverage domain adaptation, we adopt a pre-trained transformer-based model. While transformer-based models have been successfully applied for a wide range of natural language processing tasks  \cite{vaswani2017attention}, it has not been deeply studied for the QFAS task. To our knowledge, our work is among the first studies that explores the effectiveness of domain adaptation for query focused abstractive text summarization based on pre-trained transformers.}
With extensive experiments. we show that transfer learning from pre-trained transformer-based generic summarization models can be effectively utilized to tackle the few-shot learning issue in the QFAS task for both single and multi-document scenarios as well as the computational complexity related problem in long text sequences. More concretely, our contributions presented in this paper are listed below.

\introductiontable

\begin{itemize}
   
    \item \change{To address the lack of large training datasets, we propose a domain adaptation technique that utilizes 
    transfer learning via leveraging the available large generic text summarization datasets by first pre-training a transformer-based model on such datasets and then fine-tune the pre-trained model for the QFAS task by incorporating query relevance.}
    
    \item \change{To address the computational complexity problem while training neural models on multiple documents at once \cite{liuemnlpbertsum,beltagy2020longformer,choromanski2020rethinkingperformer,2020reformer,zaheer2020bigbird}, we again utilize transfer learning from pre-trained transformers using the following two novel techniques:}
    
    \begin{itemize}

    \item \change{First, we propose a weakly supervised learning model that generates the weak reference summary of each document in a document set. We then fine-tune the pre-trained transformer-based summarization model iteratively on each document for generating the query focused abstractive summary.}

    \item \change{Second, instead of generating the weak reference summary for each individual document, we propose a sentence filtering approach that selects the sentences in the document set that are most relevant to the query and feed them to the pre-trained abstractive summarization model. For this approach, we also propose a novel sequential fine-tuning technique that effectively utilizes all the gold reference summaries to provide supervised training.}

 \end{itemize}

    \item \change{We conduct comprehensive experiments with extensive ablation studies and case studies to validate our design choices on six datasets: three datasets for single-document scenarios and three datasets for multi-document scenarios. Experimental results show that our proposed approaches set new state-of-the-art results in terms of several automatic and human evaluation metrics across benchmark datasets for the QFAS task.} 
\end{itemize}

In addition to demonstrating the effectiveness of our approach, our experimental findings reveal several important new insights:  \textit{(i)} most queries do not have any relations with the input documents in the existing single-document QFAS dataset Debatepedia, \textit{(ii)} the type of attention mechanisms in the encoder can influence the performance for the QFAS task, \textit{(iii)} the domain adaptation from generic abstractive summarization models can be effective on other related tasks (i.e., abstractive answer generation in the MS-MARCO dataset), and finally \textit{(iv)} some recent transformer architectures (e.g. \citet{t5,zhang2019pegasus}) provide superior performance over their counterparts  while being utilized within our proposed approach.
As a secondary contribution, we will make our source code publicly available here: \url{https://github.com/tahmedge/PreQFAS}, so that other researchers can reproduce our experimental results and also use our codebase to push the state-of-the-art in the future.

We organize the remaining sections of this paper as follows: in Section \ref{section:rw}, we discuss the prior work on the abstractive text summarization task where we first briefly review this task for the generic abstractive text summarization scenario followed by reviewing the prior work 
where the query relevance was also taken into account. In Section \ref{section:opa}, we describe our proposed approaches for the QFAS task for both single-document and multi-document scenarios.
In Section \ref{section:experiments}, we present the datasets used in our experiments and the details of our experimental settings. The analyses of the experimental results are then presented in Section \ref{section:results}. 
Finally, we summarize our contributions with future directions in Section \ref{section:conclusion}. 

\section{Related Work} \label{section:rw}

In this section, 
we first briefly introduce readers to the generic abstractive summarization task. Then, we discuss the query focused abstractive summarization task for single-document scenarios, followed by discussing for multi-document scenarios.

\subsection{Generic Abstractive Text Summarization}
\label{section:rw_gas}

In recent years, the impressive success of neural models for sequence to sequence modeling in different natural language generation tasks \cite{nlpsurvey} has  inspired researchers to utilize the neural encoder-decoder architecture for the abstractive summary generation problem~\cite{rush,nallapati,chopra2016abstractive}. However, one major issue with the neural models for abstractive summarization is that while generating the summaries, such models tend to repeat the same word multiple times that lead to the generation of non-cohesive summaries \cite{see}. To address this issue, 
\citet{see} proposed the Pointer Generation Network (PGN) that utilized a novel copy and coverage mechanism to discourage the repetition of the same words. More recently, the BERTSUM \cite{liuemnlpbertsum} model was proposed which used the BERT model \cite{devlin2018bert} as the encoder and the decoder of the transformer model \cite{vaswani2017attention} as the decoder. The BERTSUM model that utilized fine-tuning of pre-trained transformer encoders \cite{devlin2018bert,liu2019improvingmtdnn,liu2019roberta,lan2019albert,clark2020electra,fu2021improving} showed impressive performance for the abstractive summarization task and set new state-of-the-art results in several datasets by outperforming previous neural models that leveraged the recurrent neural network architecture \cite{sequencetosequence}. The successful utilization of the transformer architecture \cite{liuemnlpbertsum} for abstractive summarization has also led to the development of more new state-of-the-art neural models that utilized this architecture for such tasks \cite{zhang2019pegasus,unilm,bart,t5,2020reformer,song2019mass,beltagy2020longformer,zaheer2020bigbird,qi2020prophetnet,fabbri2021improving}. These findings have motivated us to adopt the transformer architecture in our query focused summarization models. 

\subsection{Single-Document Query Focused Abstractive Text Summarization}
\label{section:rw_sdqfas}

While significant research has utilized neural models for the generic abstractive summarization task, applying the neural network architecture for such tasks when the query relevance is also taken into account has been rare \cite{baumel2018query}. One notable exception on utilizing neural models for such tasks is the Diversity Driven Attention (DDA) model~\cite{nema}. This model generates query focused abstractive summaries by focusing on different portions of a document based on the given query at different times. However, a key challenge in addressing the Single-Document Query Focused Abstractive  Summarization (SD-QFAS) task using neural models is that the number of datasets available for this task is quite small \cite{baumel2018query,nema,abdullah2020towardsinlg}.
To the best of our knowledge, the only available dataset for this task is the Debatepedia dataset\footnote{http://www.debatepedia.org/}, but the size of this dataset is very small compared to the datasets used for generic abstractive summarization \cite{baumel2018query,liuemnlpbertsum,see}. Thus, the lack of large training data for the SD-QFAS task in the available dataset makes this task a few-shot learning problem. To address this issue, the Relevance Sensitive Attention (RSA) for Query Focused Summarization \cite{baumel2018query} utilized transfer learning by first pre-training the PGN model \cite{see} on a large generic abstractive summarization dataset and then utilized the pre-trained model for the QFAS task to generate the summaries in the Debatepedia dataset. They found that utilizing transfer learning to be quite effective for the SD-QFAS task in that dataset.
More recently, newer models based on the recurrent neural network architecture \cite{sequencetosequence} that did not utilize transfer learning failed to outperform the RSA model in terms of different ROUGE scores \cite{canaiqfas,ecirqfas}. This may indicate that the utilization of transfer learning to tackle the few-shot learning problem has a strong effect on performance improvement in the Debatepedia dataset. However, one major limitation of the RSA model is that this model provided a poor Precision score by generating summaries much longer than the gold summaries \cite{baumel2018query}. Also, the authors did not fine-tune the pre-trained RSA model on the target dataset. In contrast, we investigate the effectiveness of fine-tuning the transformer architecture for the SD-QFAS task, motivated by the findings that fine-tuning pre-trained transformer models improve performance in a wide range of tasks including text summarization \cite{devlin2018bert,liuemnlpbertsum,2020prenlpsurvey}.

\subsection{Multi-document Query Focused Abstractive Text Summarization}
\label{section:rw_mdqfas}

The topic of query focused abstractive summarization has  remained underexplored for the multi-document scenario as well \cite{kulkarni2020aquamusegoogle}. More importantly, the currently available query focused multi-document abstractive summarization (MD-QFAS) datasets (e.g., DUC\footnote{https://www-nlpir.nist.gov/projects/duc/data.html} 2005, 2006, 2007)  do not contain any labeled training data, i.e., these datasets only provide test data \cite{baumel2018query,goodwin2020flightpegasus,su2020cairecovid,xu2021text}. 
To tackle the lack of training data for the MD-QFAS task, most previous works were based on various unsupervised approaches that could only generate extractive summaries \cite{bi-plsa,hiersum,multimr,spopt,qode,ctsum,queryfocusedsummarization2016unsupervised,queryfocusedsummarization2017unsupervised}. To generate the abstractive summaries in such tasks, \citet{baumel2018query} proposed a transfer learning technique that addressed the issue of no dedicated training data for the datasets available for such tasks. They adopted the Pointer Generation Network (PGN) \cite{see} pre-trained for the generic abstractive summarization task in a large dataset to predict the query focused summaries in the target dataset by modifying the attention mechanism of the PGN model. However, their model failed to outperform the extractive approaches in terms of various ROUGE scores. 

Here, utilizing the state-of-the-art neural summarization models \cite{liuemnlpbertsum, bart, t5, zhang2019pegasus} that leveraged supervised training is not applicable in these datasets due to the unavailability of the training data. While some recent works utilized datasets similar to the target dataset as the training set to provide supervised training \cite{li2019abstractivetkde}, some other works used similar datasets as the development dataset for hyperparameter optimization \cite{xulapata2020abstractivenew,xulapata2020coarseemnlp,su2021improveqfas}. However, while using datasets similar to the target dataset as the training data (e.g. using two DUC datasets for training the other DUC dataset), we find that these datasets only contain multi-document gold reference summaries. Thus, the state-of-the-art neural summarization models cannot be trained on such datasets since these models cannot consider long text sequences (i.e., multiple documents) as input at once due to the computational complexities \cite{zaheer2020bigbird,beltagy2020longformer}. For this reason, we utilize distant supervision from pre-trained transformers to generate the weak reference summary of each document in a document set so that the computational complexities in the MD-QFAS task can be avoided by iteratively training our model on each individual document.

Another key challenge in the MD-QFAS task is that the model needs to identify
sentences from multiple documents that are relevant to the query~\cite{msmarco}. 
Nonetheless, there could be several irrelevant sentences in different documents that are semantically similar to the relevant ones as well as to the query \cite{baumel2018query,queryfocusedsummarization2017unsupervised}, making the task of finding relevant sentences more challenging.
To identify the sentences that are relevant to the query, various approaches such as similar word count \cite{baumel2018query} or Cross-Entropy Method \cite{queryfocusedsummarization2017unsupervised} were utilized. Though neural models based on supervised training have significantly outperformed various non-neural models for the answer sentence selection task \cite{tanda2019,emnlp_bert}, due to the absence of labeled data for the relevant sentences in the MD-QFAS datasets, neural models have not been effectively utilized yet. 
Recently, \citet{tanda2019} showed that neural models such as BERT or RoBERTa pre-trained on a large question answering dataset could effectively select answers in other similar datasets without any supervised training. More recently, such pre-trained answer sentence selection models were used by \citet{xulapata2020coarseemnlp} for the MD-QFAS task. In their work, they utilized distant supervision from various question answering datasets using the fine-tuned BERT \cite{devlin2018bert} model to filter out the irrelevant sentences from the documents. However,  \citet{baumel2018query} found that filtering sentences from the input document(s) as an early step to train recurrent neural network models for query focused abstractive summarization could lead to performance deterioration. 
Thus, we also investigate how to effectively utilize sentence filtering with the pre-trained transformer models for the MD-QFAS task.  

\section{Our Proposed Approach}
\label{section:opa}

\canaifigure
Let us assume that we have a  query $Q = q_1, q_2,...,q_k$ containing $k$ words. For the QFAS task in single-document scenarios, a source document $D_{S}$ = $d_1, d_2, ...d_n$ containing $n$ words is given where the objective is to utilize the given query $Q$ to generate an abstractive summary $S = s_1, s_2,...s_m$ containing $m$ words from $D_S$. For the multi-document scenario, a set of $N$ documents $D_M$ = $D_{S_1}, D_{S_2}, ...,D_{S_N}$ are given where the goal is to generate the summary $S = s_1, s_2,...s_m$ containing $m$ words from the document set $D_M$ based on the given query $Q$.

Recall that in this paper, we aim to develop a QFAS system that can leverage the effectiveness of the transformer model \cite{vaswani2017attention,liuemnlpbertsum} to generate high quality summaries. To achieve this goal, we need to address issues such as the lack of large training datasets for the QFAS task in both single and multi-document scenarios \cite{nema,baumel2018query,queryfocusedsummarization2017unsupervised}, as well as the computational complexity related problems that occur while training transformer models in long text sequences \cite{2020reformer,beltagy2020longformer,zaheer2020bigbird,choromanski2020rethinkingperformer}. In our proposed method, we  utilize transfer learning from generic abstractive summarization models to address these issues. We choose such models for transfer learning because the available generic text summarization datasets are much larger in size compared to the QFAS datasets \cite{baumel2018query,nema}. Thus, we hypothesize that once the transformer-based models are pre-trained on large generic summarization datasets, utilizing domain adaptation from such pre-trained models on QFAS datasets would be beneficial for few-shot learning. Later on, we again utilize transfer learning from pre-trained transformers to handle the computational complexities in long sequences. Below, we describe our proposed model, denoted as: \textbf{PreQFAS}, that utilizes \textbf{Pre}-Trained Transformers for the \textbf{Q}uery \textbf{F}ocused \textbf{A}bstractive Text \textbf{S}ummarization task in detail.

\subsection{The PreQFAS model for the SD-QFAS task}
\label{section:preqfas_fsl}

For our proposed PreQFAS model, we first adopt a transformer-based \cite{vaswani2017attention} model that has been pre-trained on a large generic abstractive text summarization dataset. For that purpose, we adopt the pre-trained BERTSUM \cite{liuemnlpbertsum} model as our base model. We choose BERTSUM for three main reasons: \textit{(i) }this model achieves impressive performance for abstractive summary generation \cite{liuemnlpbertsum}, \textit{(ii)} the transformer architecture used by this model is conceptually much simpler than other recently proposed transformer-based summarization models (e.g., requires fewer number of parameters) \cite{bart,t5,zhang2019pegasus}, \textit{(iii)} this model also does not require the tuning of too many hyper-parameters to achieve the optimized performance \cite{liuemnlpbertsum}. 

Note that the BERTSUM model follows an encoder-decoder architecture that uses the BERT model as its encoder and the decoder of Transformer as its decoder. Since the BERTSUM model was designed for the generic text summarization task without considering any query relevance \cite{liuemnlpbertsum}, we  incorporate the query relevance by concatenating the query with the input document and feed into the pre-trained BERTSUM. 
Then, we fine-tune the pre-trained BERTSUM model to generate the summaries in the target QFAS dataset. More specifically, our proposed PreQFAS model performs the QFAS task in following two steps (see Figure \ref{fig1}). In the first step, we pre-train the BERTSUM model on a large training corpus of generic abstractive summarization. Then, we fine-tune the pre-trained model for the QFAS task by incorporating the query relevance. Below, we describe these two steps in detail.

\textbf{\textit{(i) Pre-training the BERTSUM Model:}} In this step, we pre-train the BERTSUM model on a large generic abstractive summarization dataset. During this pre-training stage, the model utilizes the pre-trained BERT model \cite{devlin2018bert} as the encoder and the randomly initialized transformer decoder \cite{vaswani2017attention} as the decoder.
\change{Note that this model is first trained for extractive summarization and then it is re-trained for abstractive summarization. However, unlike the original BERT model which inserts the special token \texttt{[CLS]} at the beginning of only the first sentence, the BERTSUM model \cite{liuemnlpbertsum} inserts the \texttt{[CLS]} token at the beginning of each sentence. BERTSUM does so to calculate the probability of each sentence to identify the most relevant sentences. Moreover, each sentence-pair in BERTSUM is separated by the \texttt{[SEP]} token.}
\attention

\textbf{\textit{(ii) Incorporating Query Relevance and Fine-tuning BERTSUM:}} In this step, we fine-tune the BERTSUM model on the target QFAS dataset that was pre-trained on a generic abstractive summarization dataset in the previous step. During fine-tuning, we incorporate the query relevance via concatenating the query with the document as the input to the encoder (see Figure \ref{fig1}b). We do this because we find that a similar approach of concatenating the question with the document works well with neural models for different question-answering tasks~\cite{bart}. In this way, we fine-tune a pre-trained generic abstractive summarization model for query focused abstractive summary generation to tackle the few-shot learning problem.

\textbf{Attention Mechanisms:} To utilize the query relevance in the pre-trained BERTSUM model for summary generation, we use two types of attention mechanisms (as shown in Figure \ref{fig:AttentionModelOverview}). 
 They are: (i) the bidirectional self-attention mechanism, and (ii) the query-document attention mechanism. 
 Below, we describe these two attention mechanisms.

\textit{(i) The Bidirectional Self-Attention Mechanism:} In the original BERTSUM architecture, the bidirectional self-attention mechanism \cite{devlin2018bert} is utilized by the BERT encoder to generate the encoded representation of the input text. In the bidirectional self-attention mechanism, when a pair of sentences are combined together and given as input to the BERT encoder, both sentences will give attention to each other. Thus, when we utilize the bidirectional self-attention mechanism (see Figure \ref{fig:AttentionModelOverview}a) in the PreQFAS model, both the query and the document will not only give attention to themselves, but also they will give attention to each other to provide the encoded representation of the concatenated input. 

\textit{(ii) The Query-Document Attention Mechanism:}  \citet{unilm} proposed the sequence-to-sequence language modeling objective for text sequences that are consisted of two segments. In such text sequences, each token in the first segment can only attend to the tokens in both directions within the same segment but cannot attend to any tokens in the second segment, while the tokens in the second segment can attend to the leftward tokens in their own segment as well as to all tokens in the first segment. Following this approach, we propose the Query-Document (QD) attention mechanism, where each token in the query can only attend to the tokens which are within the query while the tokens in the document can attend to all tokens in both the query and the document bidirectionally. The intuition  here is that  in the original PreQFAS model, the bidirectional self-attention allows the query to also attend to the document and thus the query segment might get influenced by the document segment. As a consequence, the final encoded representation of the concatenated input may lose some query related information and the decoder may produce summaries that may not be fully relevant to the query. 

To avoid such scenarios, we allow the query segment to only attend to itself whereas the document segment is allowed to provide a query focused representation by attending to both the query and to itself. Given the query, key, and value vectors \textbf{Q}, \textbf{K}, and \textbf{V} respectively, with ${\textbf{d}_k}$ as the square root of the dimension of \textbf{K}, we calculate the encoded representation \textbf{Z} using QD attention by adding the mask matrix $M$ in the self-attention formula of the transformer encoder \cite{vaswani2017attention}:  

\begin{equation}\label{eqqd}
    \textnormal{Z} = softmax\left(\frac{\textnormal{Q} \times \textnormal{K}^{\textnormal{T}}}{\sqrt{\textnormal{d}_k}}+{M}\right)\textnormal{V}
\end{equation}
In equation (\ref{eqqd}), ${M}_{{ij}}$ = $0$ allows attention from token $i$ to token $j$, whereas ${M}_{{ij}}$ = $-\infty$ prevents attention from token $i$ to token $j$.

\subsection{Extending PreQFAS for Long Sequences in the MD-QFAS task}
\label{section:preqfas_cc}

In this section, we discuss how we utilize our proposed PreQFAS model to address the computational complexity issue that occurs while training transformer models in long text sequences (e.g. multiple documents). 

Since the available MD-QFAS datasets only contain the gold reference summaries written for the whole document-set by human experts without containing the gold reference summary of each individual document \cite{baumel2018query,queryfocusedsummarization2016unsupervised,queryfocusedsummarization2017unsupervised}, 
neural models are ideally required to be trained on all documents in a multi-document set at once to leverage supervised training.
Nonetheless, forcing neural models to be trained on all documents at once will result in computational complexity related problems
\cite{wang2020linformer,choromanski2020rethinkingperformer,2020reformer,zaheer2020bigbird,tay2020efficient}. 

To address these issues, we propose two approaches that leverage pre-trained transformer-based models. In one approach, \change{we propose a weakly supervised learning technique that first generates the weak reference summary of each individual document in a document set. Then, we fine-tune the pre-trained transformer-based summarization model on each individual document using the weak reference summaries. In this way, we generate the summary of each individual document and then select the most relevant sentences as the final summary using a transformer-based answer selection model  \cite{laskar-LREC,laskar2020utilizing}. In another approach, instead of training our model on each document, we again utilize a transformer-based answer selection model and construct a filtered input document via selecting the sentences (up-to $n$ tokens) in the document set that are most relevant to the query. Afterward, we fine-tune the summarization model on the filtered input document. Note that we study the sentence filtering technique by applying it differently in these two approaches: for the weakly supervised learning approach, we apply it at the final stage to select the relevant sentences from the generated summary; whereas for the other approach, we apply it at the beginning to select the relevant sentences from the multi-document set.} In the following, we describe these two approaches in detail.

\distantfigureroberta
\subsubsection{\textbf{Approach 1: Weakly Supervised Learning with Distant Supervision}} 
\label{section:wsl_ds}

Figure \ref{fig:ModelOverview} shows an overview of our proposed approach that leverages weakly supervised learning. At first we generate the weak reference summary of each document in a document set by leveraging distant supervision from the multi-document gold reference summaries. Then, we propose an iterative approach that generates the query focused abstractive summary of each document by fine-tuning a pre-trained single-document generic abstractive summarization model. \change{Finally, we select the sentences (up-to $n$ tokens) that are most relevant to the query from the generated query focused summary of the multi-document set by utilizing a pre-trained answer selection model. Note that contrary to the prior work where sentence filtering was applied as an early step to filter the input document \cite{baumel2018query,xulapata2020coarseemnlp}, in this approach we apply sentence filtering during the final step to filter the generated summary.} In the following, we describe our proposed weakly supervised learning approach that tackles the computational complexity issue in the MD-QFAS task. First, we discuss how we utilize distant supervision to generate the weak reference summary of each individual document in a document set. Then, we discuss our proposed iterative approach that generates the query focused abstractive summary of each document in a document set. Finally, we describe how we select the most relevant sentences from the generated query focused summary as the final summary. 

\paragraph{\textbf{(a) Weak Reference Summary Generation:}}

We generate the weakly supervised reference summary of each document in a document set in two steps (see Figure \ref{fig:ModelOverview}a). In the first step, we utilize a pre-trained model to generate the initial weak reference summary of each document. In the second step, we replace each sentence in the generated weak reference summary by each sentence in the multi-document gold reference summaries by utilizing the RoBERTa model \cite{liu2019roberta} fine-tuned for sentence similarity modeling. For that purpose, we measure the similarity between each sentence in the multi-document gold reference summaries with each sentence in the generated weak reference summary. Then, based on the similarity score, we select the most relevant sentences from the gold reference summaries as the final weak reference summary for each document. \change{We generate the initial weak extractive reference summaries instead of directly generating the weak abstractive reference summaries since this additional step allows us to only compare the similarity between each sentence in the multi-document gold reference summaries with each sentence in the initial weak extractive reference summary. Thus, it helps our model to be more efficient during the weakly supervised reference summary generation stage by avoiding the comparison between each sentence in the multi-document set with each sentence in the gold reference summaries (see Appendix E for details).} Below, we describe these two steps in details:

\begin{itemize}

\item \textbf{Initial Weak Reference Summary Generator:}
To generate the initial weak reference summary of each document in a document set, we utilize a pre-trained transformer encoder model to generate the extractive summary of each document. To achieve our goal, we first adopt the pre-trained RoBERTa model \cite{liu2019roberta} and fine-tune it on the QA-ALL dataset of MS-MARCO \cite{msmarco} for the passage ranking (i.e., answer sentence selection) task. We choose RoBERTa in this regard due to its impressive performance on similar tasks in different answer selection datasets \cite{laskar-LREC}. Afterward, we utilize the fine-tuned RoBERTa model to measure the similarity score $C$ between the given query $Q_i$ and each sentence $S_j$ in each document $d_k$. Based on the similarity score, we select the top $3$ most relevant sentences as the weak extractive reference summary, because extracting only $3$ sentences was found effective in different extractive summarizers such as the LEAD-3 baseline as well as the BERTSUM\textsubscript{EXT} model \cite{liuemnlpbertsum}. 

\item \textbf{Final Weak Reference Summary Generator:}  The weak reference summaries generated in the previous step are extractive, while our goal is to generate abstractive summaries. Thus, we further provide distant supervision to manipulate the weak extractive reference summary generated in the previous step by replacing each sentence in the weak extractive reference summary with the most similar sentence found in the multi-document gold reference summaries written by humans. For this purpose, at first we adopt the RoBERTa model fine-tuned for the sentence similarity modeling task in the MRPC dataset \cite{liu2019roberta}. Then, for each document $d_k$ in a document set $D_i$, we utilize the fine-tuned RoBERTa\textsubscript{MRPC} model to measure the similarity between each sentence $S_j$ in the weak extractive reference summary and each sentence $S_g$ in the gold reference summaries. Based on the similarity score, each sentence in the weak extractive reference summary of a document is replaced with the most relevant sentence found in the multi-document abstractive gold reference summaries. Note that for a document $d_k$ when a sentence $S_g$ from the gold reference summaries is already used to replace a sentence $S_j$ in the weak extractive reference summary, then for the same document $d_k$ we do not consider the sentence $S_g$ again for replacement. Instead, we use the next most relevant sentence from the multi-document gold reference summaries for replacement. The resulting summaries generated in this step can be considered as weak abstractive reference summaries because they are constructed from the gold reference summaries which are written by human annotators. In the following, we discuss how we train our model using these weak abstractive reference summaries.
\end{itemize}

\paragraph{\textbf{(b) Iterative Fine-Tuning:}}
 In the MD-QFAS task, since the available datasets are also small in size \cite{queryfocusedsummarization2017unsupervised,xulapata2020coarseemnlp,dualces}, we again utilize the PreQFAS model proposed in Section \ref{section:preqfas_fsl} to address the few-shot learning problem. However, the PreQFAS model is based on the BERTSUM model that was pre-trained for the single-document generic summarization task by considering at most 512 tokens \cite{liuemnlpbertsum}. In reality, the total number of tokens in a document set in multi-document scenarios could be much larger than 512 tokens \cite{baumel2018query,queryfocusedsummarization2017unsupervised}. \change{Thus, to avoid the computational complexities of training transformer-based models in such long sequences at once \cite{2020reformer,beltagy2020longformer,zaheer2020bigbird,choromanski2020rethinkingperformer}, we take an iterative approach where we fine-tune the pre-trained summarization model on each individual document in a multi-document set (see Figure \ref{fig:ModelOverview}b).} In this approach, similar to the PreQFAS model proposed in Section \ref{section:preqfas_fsl} for the SD-QFAS task, we first adopt the pre-trained BERTSUM model. Then, we incorporate the query relevance into the pre-trained BERTSUM and fine-tune it using the weak abstractive reference summary to generate the query focused abstractive summary of each document in the given document set. Finally, we select the top $N$ most relevant sentences from the generated summaries as the final summary. We describe the final summary selection procedure in detail next.

\paragraph{\textbf{(c) Summary Sentence Selection:}} In this stage, for each document set, all the sentences in the query focused abstractive summaries generated in the previous step are ranked using a fine-tuned RoBERTa model. For this purpose, we adopt the RoBERTa model fine-tuned for the answer selection task in the MS-MARCO dataset, which we also utilized for initial weak reference summary generation. The fine-tuned RoBERTa\textsubscript{MS-MARCO} model is then utilized to measure the relevance between each sentence $S_i$ in the generated summary and the query $Q_j$ for the document set $D_j$ to select the sentences that are most relevant to the query as the final summary. \change{In this way, we utilize sentence filtering in the final step such that the total length of the selected sentences in the final summary does not exceed $n$ tokens (see Figure \ref{fig:ModelOverview}c).} To reduce redundancy in the final summary, we use the Trigram Blocking \cite{paulus2018deepICLR}.

\subsubsection{\textbf{Approach 2: Sequential Fine-Tuning with Sentence Filtering}}
\label{section:sft_sf}

\change{In our weakly supervised learning approach demonstrated earlier, we apply sentence filtering in the final stage to identify the most relevant sentences in the generated summary. While most summarization models have attempted sentence filtering in the early stage where the irrelevant sentences or paragraphs were filtered out from the source document(s) prior to generating the abstractive summaries, the results obtained by these models were conflicting \cite{liu2019hierarchical, baumel2018query}. For instance,  \citet{liu2019hierarchical} found that sentence filtering as an early step did not deteriorate the performance in the generic abstractive summarization task, whereas \citet{baumel2018query} found that such a step deteriorated the performance in  query-based multi-document abstractive summarization. To investigate the performance of sentence filtering as an early step, we develop the following approach. First, we select the sentences from the multi-document set that are most relevant to the query to construct a filtered input document. Then, we give the filtered input document as input to a PreQFAS model for fine-tuning. 

}

More specifically, we first adopt a transformer-based answer selection model to identify the sentences in a document set that are most relevant to the query. Then, based on the relevance score, we rank the sentences in the document set. Next, we keep selecting the sentences until the total length of the selected sentences along with the query does not exceed $n$ tokens. In this way, we create a filtered input document. Then, we utilize our proposed PreQFAS architecture to combine the query and the filtered input document together in order to give them as input to the BERTSUM model pre-trained for generic abstractive summarization. 
We then propose a sequential technique to fine-tune the BERTSUM model to provide supervised training via leveraging all gold reference summaries written by different human annotators for a given document set to generate the query focused abstractive summary. The overall approach is shown in Figure \ref{filterfig}. Below, we describe our input document filtering process followed by the summary generation process in detail.

\mdqfasfilterfigure
\mdqfasfilterfigureexample
\paragraph{\textbf{(a) Sentence Filtering:}} In this step, for each document set we measure the relevance of all sentences to the given query $Q_i$.
\change{For that purpose, we adopt the RoBERTa model and fine-tune it for the answer ranking task in the QA-ALL dataset of MS-MARCO \cite{liu2019roberta,msmarco,laskar-LREC}. Based on the relevance score, we then rank all the sentences in a document set. Afterward, we concatenate the query and the ranked sentences and consider the first $n$ tokens as our input document for the summarization model. Note that this sentence filtering approach not only allows us to leverage the state-of-the-art neural summarization models to provide supervised training for the MD-QFAS task, but also allows us to overcome the computational complexity issue that occurs while training neural models on long documents.}

\paragraph{\textbf{(b) Sequential Fine-Tuning:}} In the previous step, we 
select the most relevant sentences to the query $Q_i$ to construct the input document containing $n$ tokens. In this way, the multiple documents are converted into a single-document that consists of only those sentences that are most relevant to the query. Therefore, the filtered document allows us to leverage the effectiveness of fine-tuning pre-trained single-document generic abstractive summarization models. Thus, we adopt the BERTSUM model that was pre-trained for single-document abstractive summarization and fine-tune it to generate the query focused abstractive summary for the given input document (i.e., the filtered input document constructed from a given document set).

    \begin{figure*}[t!]
        \begin{center}
        \includegraphics[width=\linewidth]{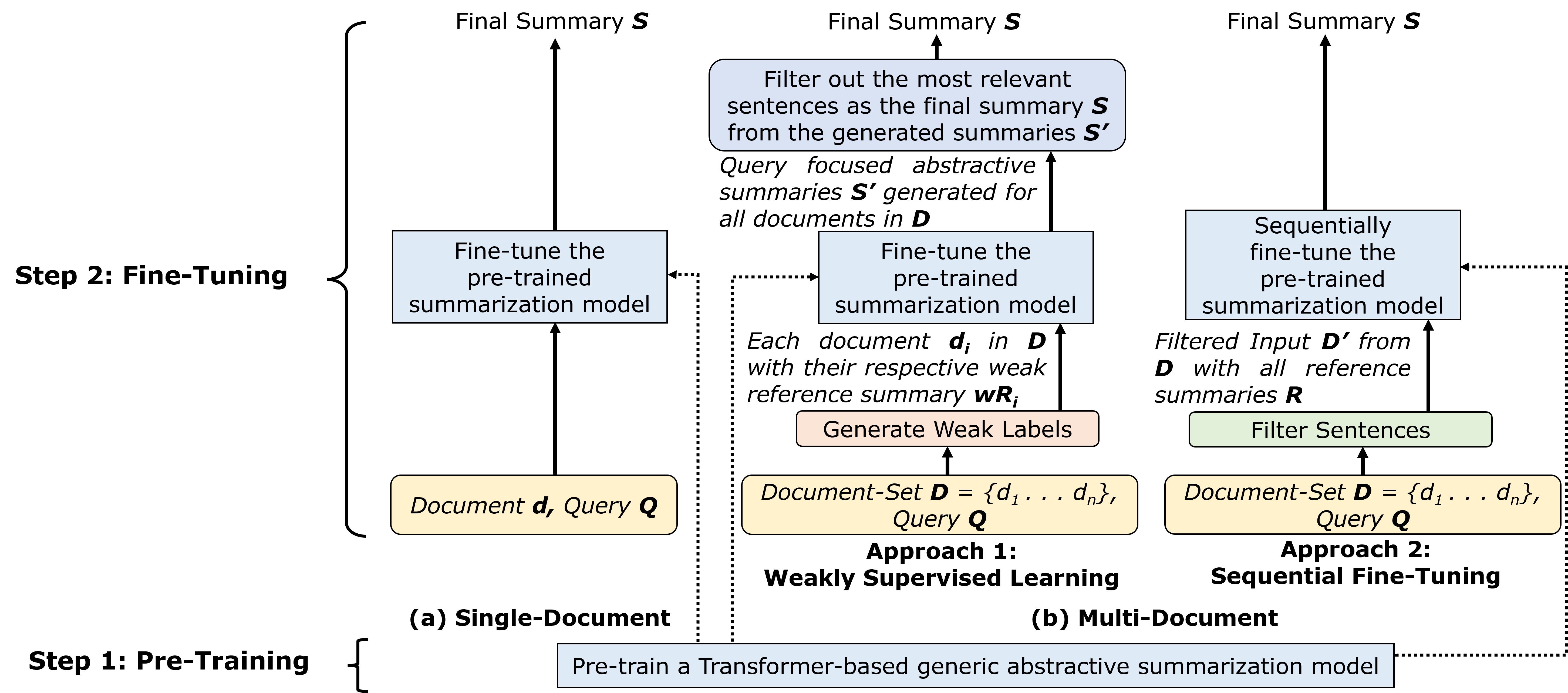}
        \caption{
        {An overview summary of our proposed approaches: (a) one approach for single document scenarios, (b) two approaches for multi-document scenarios.}
        }
     
        \label{figureoverallarchitecture}
           \end{center}
    \end{figure*}

Since the available MD-QFAS datasets contain multiple gold reference summaries written by different human experts for the same query \cite{queryfocusedsummarization2017unsupervised}, training neural models using multiple gold reference summaries will allow an encoder-decoder model to enhance its vocabulary in the decoder \cite{rush,nallapati}. In order to leverage the advantage of multiple gold summaries, we propose a sequential fine-tuning model. In our proposed approach, if there are $k$ gold summaries for a given training document set, then we fine-tune the model $k$ times where each fine-tuning run\footnote{Each fine-tuning run may consist of $X$ epochs or $Y$ steps used for training a neural network model.} will have gold summaries different than the other runs for the same filtered input document. Thus, in Figure \ref{filterfig}(c), the BERTSUM model will be fine-tuned $k$ times. Note that for the first fine-tuning run, we adopt the model for fine-tuning that is pre-trained on a generic abstractive summarization task. For the subsequent runs, we fine-tune the model that is fine-tuned in the immediate previous run. We show the sequential fine-tuning process in Figure \ref{filterexamplefig}.

\change{
\subsection{Summary of the Proposed Models}

So far, we have presented three different approaches for the QFAS task, one for the single-document scenario and two for the multi-document scenario (see Figure~\ref{figureoverallarchitecture}). In all three approaches, we first pre-trained a transformer-based summarization model (e.g., BERTSUM) on a large generaic abstraction summization dataset.  For the single-document scenario (see Figure~\ref{figureoverallarchitecture}a), we fine-tune the pre-trained model on the target query focused summarization dataset by incorporating query relevance. For the multi-document scenario (see Figure~\ref{figureoverallarchitecture}b), we propose two approaches: \textit{(i)} PreQFAS\textsubscript{WSL}: a weakly supervised approach that generates weak labels, i.e., the weak reference summary of each document in a document set so that we can fine-tune a pre-trained transformer-based summarization model by avoiding the computational issues,\textit{ (ii)} PreQFAS\textsubscript{SFT}: a sequential fine-tuning technique that first selects the most relevant  sentences from the multi-document set and sends them to a pre-trained transformer-based summarization model to fine-tune the model sequentially using multiple gold reference summaries. In the first approach, we perform the sentence filtering at the last step of summary generation, while in the second approach we perform the sentence filtering as an early step.}

\section{Experimental Setup}
\label{section:experiments}

In this section, we describe the datasets that we use to evaluate the effectiveness of our approach, followed by the evaluation metrics, the training parameters that have been used in our experiments, and finally the implementation details of our proposed models.

\subsection{\textbf{Datasets}} 
\label{section:datasets}

For the QFAS task in single-document scenarios, we primarily use the Debatepedia \cite{nema} dataset to evaluate our proposed approach. Additionally, due to the lack of available datasets for this task, we modify the QA-NLG dataset from MS-MARCO \cite{msmarco} and utilize it for the SD-QFAS task in order to investigate the generalized effectiveness of our proposed approach across different datasets in a related domain. For the QFAS task in multi-document scenarios, we use three datasets from DUC (2005, 2006, 2007) since these datasets are widely used for such tasks. Below, we discuss all datasets in detail. 

\textbf{Debatepedia:} Debatepedia is an encyclopedia of pro and con arguments and quotes on debate topics. \citet{nema} utilized Debatepedia to create a dataset containing 13,573 instances for the SD-QFAS task. The average number of words per document, summary, and query in the Debatepedia dataset is 66.4, 11.16, and 9.97 respectively. They used 10-fold cross-validation in their experiments, where each fold has 80\% data for training, 10\% for validation, and 10\% for testing which resulted in average instances of 10,859 for training, 1,357 for testing, and 1,357 for validation respectively. We pre-processed the dataset by removing the start token  \texttt{$<$s$>$} and the end token \texttt{$<$eos$>$}.

{\textbf{MS-MARCO:}} As mentioned earlier, due to the lack of datasets for the SD-QFAS task, we also utilize the QA-NLG dataset from MS-MARCO \cite{msmarco} which was designed for the abstractive answer generation task. With 1,53,725 training samples, this dataset is much larger than the Debatepedia dataset. Therefore, this dataset can give useful insights to investigate the generalized effectiveness of our model rather than only evaluating its performance for few-shot learning.
In the original task setup, a set of passages along with a query are given and the goal is to generate an abstractive answer from the most relevant passage among them. To treat this dataset as an SD-QFAS dataset, we follow the work of \citet{nishida2019multi}, where they  utilized only the gold passages in the training set as well as in the development set in one of their experiments. We use this dataset similarly by  utilizing only the gold passage as the single source document along with the associated query. We use the development set of this dataset that contains 12,467 queries for evaluation. During experiments, we used 10\% data from the training set for validation. 

{\textbf{DUC:}} We use the DUC 2005, 2006, and 2007 datasets for the MD-QFAS task.
The number of multi-document sets were 50, 50, and 45 while the average number of documents in each multi-document set were 32, 25, and 25 in DUC 2005, 2006 and 2007 datasets respectively~\cite{queryfocusedsummarization2017unsupervised}. Each document set is associated with a topic statement (considered as the query) and the goal is to generate a summary containing at most 250 words from the document set based on that query. 
Given the absence of the training data, to evaluate our model in each year's dataset we use the datasets from other two years for training. From each year's training data, we randomly selected 20\% of the document sets for validation while the rest were used for training.

\subsection{\textbf{Evaluation Metrics}} \label{section:em} To evaluate the performance of our models in different datasets, we  select  the evaluation metrics  by following the prior works \cite{nema,baumel2018query,nishida2019multi}. For the Debatepedia dataset, we report the results based on the Recall-Oriented Understudy for Gisting Evaluation (ROUGE) \cite{rouge} metric in terms of the ROUGE-1, ROUGE-2, and ROUGE-L scores\footnote{We used the following package for calculation: https://pypi.org/project/pyrouge/}. Though the prior works that used the Debatepedia dataset reported the ROUGE scores only in terms of the Recall metric \cite{nema,baumel2018query}, we additionally included the Precision and the F1 metrics.
We calculated the result based on the average across 10-folds. 
For the MS-MARCO dataset, the prior work used the Bilingual Evaluation Understudy (BLEU) \cite{bleu,reiter2018structuredbleu} metric based on unigrams in addition to the ROUGE-L metric for performance evaluation \cite{nishida2019multi}.  We also use these two metrics in terms of the F1 score to evaluate our proposed models.  
For the DUC datasets, we reported the results based on both Recall and F1 metrics in terms of ROUGE-1, ROUGE-2, and ROUGE-SU4 scores \cite{rouge} using the standard parameter setting\footnote{ROUGE-1.5.5.pl -a -c 95 -m -n 2 -2 4 -u -p 0.5 -l 250} as used in prior work \cite{queryfocusedsummarization2017unsupervised,dualces}. 

\subsection{\textbf{Training and Parameter Settings}} 
\label{section:tps}

In order to pre-train the BERTSUM model on a generic abstractive summarization dataset, we adopt the BERTSUM models that are pre-trained either on the CNN/DailyMail (CNN-DM) dataset or the XSUM dataset, as used by  \citet{liuemnlpbertsum} for generic abstractive summarization. 

For pre-training, we kept the parameters similar to the original work \cite{liuemnlpbertsum}: dropout = 0.1, label smoothing with smoothing factor = $0.1$, the hidden units in the transformer decoder = $768$ and the hidden size for all feed-forward layers = $2048$, the warmup\_steps for the encoder = $20000$ and for the decoder = $10000$, the learning\_rate for the encoder = $0.002$ and for the decoder = $0.1$. The batch size was also set to $140$. When the CNN-DM dataset was used for pre-training, the total pre-training step was: $148000$, while for the XSUM dataset the total pre-training step was $30000$. 

To fine-tune the  BERTSUM model on the target SD-QFAS datasets, we set new values to the following parameters: batch size = 500, warmup\_steps\_encoder = $6000$, and warmup\_steps\_decoder = $2000$. We ran additional $12000$ training steps for fine-tuning to set the total training steps = $160000$ when the CNN-DM dataset was initially used for pre-training. When we used the XSUM dataset for pre-training, we ran additional $30000$ training steps (in total $60000$) to do the fine-tuning. Moreover, for Debatepedia, we truncated each input document to $100$ tokens and at most $25$ tokens for each generated summary. For MS-MARCO, each input document was truncated to $256$ tokens and each generated summary had 100 tokens. As used in the original BERTSUM model \cite{liuemnlpbertsum}, we also utilized the beam search decoding mechanism with size = $5$. 

To fine-tune the BERTSUM model for the MD-QFAS task, we kept most parameters similar to what we used for the SD-QFAS task. However, in this case, we only ran 50 additional steps from the pre-trained model for fine-tuning with batch size equal to $250$. For the RoBERTa sentence similarity model \cite{liu2019roberta}, we fine-tuned its pre-trained model for the pair-wise sentence classification task using the same parameters that were utilized by \citet{laskar-LREC}. For all tasks, 
we used the models for evaluation on the test dataset that performed the best on the validation dataset.

 \canaitable

\subsection{\textbf{Implementation}} \label{section:implementation}
For the RoBERTa model, we use its Large version \cite{liu2019roberta} for all cases for the MD-QFAS task: when we generate the initial weak reference summaries and while ranking the generated query focused abstractive summaries in the final step of the PreQFAS\textsubscript{WSL} model, as well as when we create the filtered input document for the PreQFAS\textsubscript{SFT} model. To implement this model, we use the Transformer library of HuggingFace \cite{Wolf2019HuggingFacesTS}. For the BERTSUM model, we utilize the BERTSUM\textsubscript{EXT-ABS} architecture that was used by  \citet{liuemnlpbertsum}. For implementation, we use the official source code of the BERTSUM\footnote{https://github.com/nlpyang/PreSumm} model \cite{liuemnlpbertsum}. All of our experiments were run using NVIDIA V100 with 4 GPUs.

\section{Results and Discussions} \label{section:results}

In this section, we discuss the performance of our proposed PreQFAS architecture in different datasets. We first demonstrate our findings in the SD-QFAS task, followed by discussing our findings in the MD-QFAS task. 

\subsection{Performance on the SD-QFAS task} \label{section:results_sdqfas}

In the following, we first discuss the performance of the PreQFAS model\footnote{To utilize PreQFAS for SD-QFAS, we adopt the BERTSUM model pre-traiend on the XSUM dataset.} for few-shot learning presented in Section \ref{section:preqfas_fsl} on the Debatepedia dataset, followed by on the MS-MARCO dataset. Finally, we present a set of case studies as well as ablation studies to provide a deeper understanding of the effectiveness of our approach.

\subsubsection{\textbf{Performance on the Debatepedia dataset}} \label{section:results_debatepedia}

In order to compare the performance of our proposed model, we adopt the original BERTSUM model \cite{liuemnlpbertsum} as a baseline (denoted as QR-BERTSUM\textsubscript{Vanilla}) by concatenating the query with the document as input and train it end-to-end only on the target Debatepedia dataset. In addition to this baseline, we also compare  our model with some other models that were evaluated on this dataset: the DDA model by \citet{nema} which was the first model proposed for this dataset; the RSA model \cite{baumel2018query} which provided the state-of-the-art performance among the recurrent neural network models (in terms of ROUGE-1 and ROUGE-L), the Overlap-Wind model \cite{ecirqfas} (set a new state-of-the-art based on ROUGE-2); and the recently proposed Selection Driven model \cite{canaiqfas}. For  comparisons, we evaluate our PreQFAS model for both the query-document attention and the bidirectional self-attention. 

Table~\ref{tab1} shows the results for our proposed model compared to other models. We find that the PreQFAS model with both attentions significantly improved the performance over the QR-BERTSUM\textsubscript{Vanilla} model that did not leverage any transfer learning from generic abstractive summarization datasets. This improvement suggests the effectiveness of domain adaptation from pre-trained generic abstractive summarization models for the SD-QFAS task in the Debatepedia dataset.

When we compare the performance between different attentions in the PreQFAS model, we observe that both attentions provide the exact same result in terms of most ROUGE scores with only a few exceptions. We find that the PreQFAS model with the bidirectional self-attention outperforms its QD attention counterpart in two cases in terms of the Precision metric, with an improvement of 0.17\% for both ROUGE-1 and ROUGE-L scores. The only case when The PreQFAS model with the QD attention outperforms the PreQFAS with the bidirectional self-attention is based on F1 in terms of ROUGE-1, with an improvement of 0.34\%. The overall result in the Debatepedia dataset suggests that introducing the QD attention is not more effective than the bidirectional self-attention.

In comparison to the prior work, we observe that the proposed PreQFAS model sets a new state-of-the-art result in all three ROUGE scores for both attentions. More specifically, in terms of Recall, we find that the PreQFAS model (for both attentions) has an improvement of 9.23\%, and 23.59\% in terms of ROUGE-1, and ROUGE-L respectively, over the RSA model \cite{baumel2018query}. As mentioned by \citet{baumel2018query}, the RSA model provided a very low ROUGE Precision score (the paper does not report the exact score) by generating very long summaries which are 10 times longer than the required length. In contrast, our proposed model shows a high Precision score by effectively generating summaries according to the required length. 
We also observe a huge gain over other models based on the ROUGE-2 score, with an improvement of 140.43\%, 180.75\%, 64.96\%, 48.20\%  over the DDA \cite{nema}, RSA \cite{baumel2018query}, Selection Driven \cite{canaiqfas}, and Overlap-Wind \cite{ecirqfas} models respectively in terms of the Recall metric.

\subsubsection{\textbf{Performance on the MS-MARCO dataset}} \label{section:msmarco}

\msmarcoqfastable

For the MS-MARCO dataset, in addition to the baseline\footnote{The baseline QR-BERTSUM\textsubscript{Vanilla} model in Table \ref{tab2} was trained end-to-end on the MS-MARCO dataset.}, we compare our proposed model with the MASQUE model \cite{nishida2019multi}, the current state-of-the-art in this dataset. We observe from Table \ref{tab2}  that our proposed PreQFAS model (for both attentions) again outperforms the baseline model. More specifically, our best performing PreQFAS using the bidirectional self-attention outperforms the baseline QR-BERTSUM\textsubscript{Vanilla} with an improvement of 9.50\% in terms of ROUGE-L and 14.53\% in terms of BLEU-1. 
These improvements demonstrate the effectiveness of our proposed approach that utilizes transfer learning by fine-tuning pre-trained generic abstractive summarization models. Our model with bidirectional self-attention also outperforms the MASQUE \cite{nishida2019multi} model by 2.94\% in terms of BLEU-1 while the result was almost identical in terms of ROUGE-L.

\change{A possible explanation for why our model could not outperform the MASQUE model in terms of ROUGE-L is that the size of the training data used for the MASQUE model was much larger than the dataset that we used to train our model. In our case, we utilize the QA-NLG dataset from MS-MARCO where the training data contains 1,53,725 instances whereas the MASQUE model used the QA-ALL dataset where the training data contains 8,08,731 instances \cite{msmarco}. Despite these differences, with less training data our model outperforms the MASQUE model in terms of BLEU-1 and achieves similar results in terms of ROUGE-L.}

When we compare between different attentions in the MS-MARCO dataset, we find that the QD attention is much less effective than the bidirectional self-attention mechanism. More specifically, we find that when the QD attention is used instead of using the bidirectional self-attention, the performance is deteriorated by 7.78\% in terms of ROUGE-L and 10.32\% in terms of BLEU-1. This can be explained based on the findings of \citet{peters-etal-2019-tune}, as they suggest that the performance in downstream tasks depends on the similarity between the pre-training stage and the fine-tuning stage. Since the BERT encoder was pre-trained by using the bidirectional self-attention, the utilization of the QD attention only during fine-tuning could possibly be the reason behind poorer performance. 

\ablationqfas

\zssdqfas

Interestingly, while comparing the performance of these attentions 
in different datasets, we observe a very surprising trend. Based on our experiments, we find that both the QD attention and the bidirectional self-attention perform similarly in the Debatepedia dataset (see Table \ref{tab1}), whereas the QD attention performs more poorly than the bidirectional self-attention in the MS-MARCO dataset (see Table \ref{tab2}). Furthermore, we find that when domain adaptation from pre-trained summarization models is not utilized, the performance of the baseline QR-BERTSUM\textsubscript{Vanilla} model in the MS-MARCO dataset (see Table \ref{tab1}) is much better than its performance in the Debatepedia dataset (see Table \ref{tab2}). This could be due to the fact that the total training instances (1,53,725 examples) in the MS-MARCO dataset is almost 15 times higher than the total training instances (10,859 examples) in the Debatepedia dataset. Nonetheless, the performance improvement in our PreQFAS model from the baseline in all datasets shows that our proposed model is not only effective to handle the few-shot learning problem, but also it can achieve a huge performance gain when the size of the training dataset is large.

\subsubsection{\textbf{Ablation Study}} \label{section:as_sdqfas}

In this section, we conduct ablation tests to demonstrate the effectiveness of using different components in our proposed model. For the ablation test, our key questions are 

\begin{itemize}
    \item \textbf{Why \textit{fine-tuning} helps to improve the performance?} \textit{To  answer this question, we simply use the pre-trained model for inference without fine-tuning it on the target dataset.}

    \item \textbf{To what extent utilizing the \textit{query relevance} is useful?} \textit{To answer this question, we remove the query as input to our model.}
\end{itemize}
We show the result of our ablation test in Table \ref{tab:ablationqfas}. We can readily see that removing fine-tuning  degrades the performance in both the MS-MARCO and the Debatepedia dataset \textbf{significantly} (based on paired t-test with $p$ $\leq$ $.05$ on both datasets).

Removal of query relevance also leads to huge performance deterioration in the MS-MARCO dataset, which is \textbf{statistically significant} based on paired t-test ($p$ $\leq$ $.05$). Surprisingly, we  find that the performance deterioration in Debatepedia is very small (less than 1\%) and this difference was \textbf{not statistically significant} according to the paired t-test ($p$ $>$ $.05$). Such striking difference in performance between MS-MARCO and Debatepedia suggests that the queries in the Debatepedia dataset may not be effective for summarization.

\change{
\subsubsection{\textbf{Case Studies}} \label{section:cs_sdqfas}

While we found that fine-tuning a pre-trained generic summarization model on the target domain to be effective, we now investigate if this approach is still effective for the zero-shot learning scenario where the training dataset for the target domain is not available. To answer this question, we create a zero-shot learning setup, where we first adopt the BERTSUM model (pre-trained on the XSUM dataset) and then instead of fine-tuning the model on the target dataset, we fine tune it on the MS-MARCO dataset. We choose the MS-MARCO dataset because it is much larger in size and so we hypothesize that fine-tuning on it will provide better generalization.

We study the zero-shot learning scenario with two different target datasets: (i) Debatepedia, and (ii) MEDIQA-Answer Summarization (MEDIQA-AnS) dataset \cite{savery2020questionmediqaans}. MEDIQA-AnS is a question answering dataset in the healthcare domain, where given a question and a long answer, the goal is to summarize the answer. The MEDIQA-AnS dataset is particularly suitable for the zero-shot setup because it does not contain any training data and so the model needs to generate the abstractive summaries for a given question without any in-domain knowledge. The MEDIQA-AnS dataset has two versions based on the type of the input document:  (i) Pages Version: the input is comprised of some web pages that are relevant to the question, and  (ii) Passages Version: the input only contains some passages from the relevant web pages. We use both versions in our study.

For performance comparisons, we use the BERTSUM model as a baseline that was pre-trained only on the XSUM dataset and did not leverage any fine-tuning. The result of our case study is shown in Table \ref{tab:zsqfas}. We observe that in all zero-shot learning scenarios, fine-tuning the model on a dataset that is not from the target domain is more effective than no fine-tuning at all. For instance, despite the fact that the PreQFAS model was fine-tuned on the MS-MARCO dataset that is different than the target MEDIQA/Debatepedia dataset, it still demonstrates superior performance over the baseline BERTSUM model. 

In addition to this case study, we conduct another case study where we investigate the effects of using different datasets for pre-training that can be found in Appendix A.

}

\dbpediaanalysis

\analyzedebatepedia  

\subsubsection{\textbf{Analyzing the Debatepedia dataset}} \label{section:analyze_debatepedia} \change{Due to the surprising performance in the Debatepedia dataset that we observe in our ablation study in Section \ref{section:as_sdqfas} after removing the query relevance, we manually analyze the dataset to find out the possible reasons.} For our analysis, we randomly sampled 100 query-document-summary tuples. The result of our analysis is shown in Table \ref{tab:dbpediaanalysis}.  We observe that many queries in this dataset are not relevant to the source document as well as to the reference summary (about 52\%). Table \ref{tab:analyzedebatepedia}(a) shows such an example from this dataset where the query has no relevance with the source document and the gold summary. We also find many examples where the query contains only one word, which partially explains the lack of effectiveness of incorporating the query. Furthermore, we find that most queries are just yes/no type questions (see Table \ref{tab:analyzedebatepedia}(b) for an example) that do not necessarily require to generate a query focused summary (about 70\%). Besides, we observe that among the documents where the generated summaries are relevant to the query, excluding queries from most of these documents will not have any negative effects to generate the relevant summaries. The possible reason behind this is that since the average document length in Debatepedia is very small (66.4 words on average per document), the gold summaries for most documents tend to reflect the overall generic summaries of these documents (81\% according to the result in Table \ref{tab:dbpediaanalysis}) where the queries that are used for such documents do not influence the summary (see Table \ref{tab:analyzedebatepedia}(c) for such an example). 

These findings strongly indicate that most queries in Debatepedia are not relevant to the generated summaries and as such this dataset can be considered more of a generic summarization dataset (as opposed to a query focused summarization dataset).

\subsection{Performance on the MD-QFAS task} \label{section:results_mdqfas}

\wstable

We now analyze the effectiveness of our approach\footnote{To utilize PreQFAS for MD-QFAS, we adopt the BERTSUM model pre-traiend on the CNN-DM dataset.} in the multi-document query focused summarization scenario. Recall that we proposed two model variations for this scenario in Section \ref{section:wsl_ds} and Section \ref{section:sft_sf} respectively. We denote our proposed approach that utilizes \textbf{W}eakly \textbf{S}upervised \textbf{L}earning as \textbf{PreQFAS\textsubscript{WSL}} (see Section \ref{section:wsl_ds}), while the one that utilizes \textbf{S}equential \textbf{F}ine-\textbf{T}uning as \textbf{PreQFAS\textsubscript{SFT}} (see Section \ref{section:sft_sf}).

We compare our models with  two  baselines that utilize the pre-trained BERTSUM model for zero-shot transfer learning without leveraging any supervised signals and fine-tuning. For each document, one baseline generates extractive (EXT) summary (\textbf{BERTSUM\textsubscript{EXT}}), while the other generates abstractive (ABS) summary (\textbf{BERTSUM\textsubscript{ABS}}). Similar to the  \textbf{PreQFAS\textsubscript{WSL}} model, the generated summaries in both baselines are also ranked using the RoBERTa model. In addition, we compare our models with four recent works: i) CES-50 \cite{queryfocusedsummarization2017unsupervised}, ii) RSA \cite{baumel2018query}, iii) Dual-CES \cite{dualces}, and iv) QUERYSUM \cite{xulapata2020coarseemnlp}.

\subsubsection{\textbf{Performance on DUC datasets}} \label{section:results_duc}

The results of our experiments in DUC 2005, DUC 2006, and DUC 2007 datasets are shown in Table \ref{tab:wstable}. In all three datasets, both variations of our model outperform all the prior works in terms of the F1 metric. 
More specifically, in the DUC 2005 dataset, the \textbf{PreQFAS\textsubscript{SFT}} sets a new state-of-the-art in all ROUGE scores (outperforms the previous state-of-the-art DUAL-CES 
by 6.85\%, 22.94\%, and 14.81\% in terms of ROUGE-1, ROUGE-2, and ROUGE-SU4 scores respectively). In other two datasets, \textbf{PreQFAS\textsubscript{WSL}} model also provides state-of-the-art performance across all ROUGE scores. In DUC 2006, it beats the QUERYSUM model 
by  4.54\%, 13.47\%, and 7.52\% in terms of ROUGE-1, ROUGE-2, and ROUGE-SU4 respectively. Finally, in DUC 2007, it made an improvement of 3.28\% over QUERYSUM based on ROUGE-1, while 5.60\% and 5.29\% over Dual-CES based on ROUGE-2 and ROUGE-SU4 respectively.

In terms of the Recall metric, the proposed \textbf{PreQFAS\textsubscript{WSL}} model outperforms the prior state-of-the-art \cite{dualces} in ROUGE-2 in DUC 2005 and 2006 with an improvement of 13.63\% and 6.05\% respectively. 
For other two metrics (i.e. ROUGE-1 and ROUGE-3) based on Recall, none of our models could outperform the prior state-of-the-art models in any datasets. However, the results are still comparable.

\mdqfasablationall

While comparing between the zero-shot baselines, we find that in both DUC 2005 and DUC 2006 datasets, our abstractive baseline outperforms its extractive counterpart. However, in the DUC 2007 dataset, we find that the extractive baseline performs better than the abstractive one. This may indicate that the gold reference summaries in the DUC 2007 dataset are more extractive in nature. Moreover, while comparing these two baselines with the proposed \textbf{PreQFAS\textsubscript{WSL}} model, we find that for all ROUGE scores, the performance improvement in our proposed model is \textbf{statistically significant} based on paired t-test ($p$ $\leq$ $.05$).

\subsubsection{\textbf{Ablation Study}} \label{section:as_mdqfas} 
\change{We conduct four ablation tests for the multi-document scenario. The first three tests examine how the following components of our weakly supervised approach} impact the performance of the \textbf{PreQFAS\textsubscript{WSL}} model: (i) \textit{Distant Supervision}, (ii) \textit{Trigram Blocking}, and (iii) \textit{Weakly Supervised Learning}.

Our ablation study results in Table \ref{tab:mdqfasablationall} suggest that instead of leveraging \textit{Weakly Supervised Learning} to fine-tune the BERTSUM model, if we directly rank the sentences in the source documents using the RoBERTa\textsubscript{MS-MARCO} model and select the first 250 tokens as the summary, the performance is \textbf{significantly} degraded (based on paired t-test with $p$ $\leq$ $.05$). The performance is also deteriorated if we exclude \textit{Distant Supervision} by removing the RoBERTa\textsubscript{MRPC} model as well as if the \textit{Trigram Blocking} is not utilized. However, in these two cases, the performance deterioration is \textbf{not statistically significant} based on paired t-test ($p$ $>$ $.05$).  

\change{In our final ablation test, we study the effect of sequential fine-tuning in the \textbf{PreQFAS\textsubscript{SFT}} model. For this ablation test, instead of sequential fine-tuning where we vary gold reference summaries in multiple runs, we fine-tune only once by varying the gold reference summaries when the same filtered document is given as input to the summarization model in different batches. We find that when the fine-tuning is done only once, the performance deterioration from our proposed sequential fine-tuning approach is \textbf{statistically significant}. This indicates the effectiveness of sequential fine-tuning with multiple runs where we vary the gold reference summaries in each run. }

\change{
\subsubsection{\textbf{Case Studies}} \label{section:cs_mdqfas} In this section, we perform case studies 
to investigate how modifying different stages of PreQFAS impact its performance. For these case studies, we investigate the following questions for the {PreQFAS\textsubscript{WSL}} model and the PreQFAS\textsubscript{SFT} model:

\renewcommand{\theenumi}{\roman{enumi}}%
\begin{enumerate}

    \item {\textbf{For PreQFAS\textsubscript{WSL}}, we investigate what happens if we fine-tune other pre-trained transformer-based generic abstractive summarizaiton models such as BART \cite{bart}, PEGASUS \cite{zhang2019pegasus}, and T5 \cite{t5} instead of BERTSUM?}

    \item {\textbf{For PreQFAS\textsubscript{SFT}}, we investigate how the total number of gold reference summaries $K$ used for fine-tuning impacts the performance.}
\end{enumerate}
}

\csmdqfasf

Below, we present our findings for each of the above questions. 

\paragraph{\textbf{(i) Fine-Tuning Other Models for Summary Generation using PreQFAS\textsubscript{WSL}:}} Recall that we fine-tune the pre-trained BERTSUM model\cite{liuemnlpbertsum} to generate the query focused abstractive summaries since it uses a simple transformer architecture that has less complexities (see Section \ref{section:preqfas_fsl} for more details). Thus, it allows us to evaluate the effectiveness of our proposed approach while utilizing a conceptually simple model. Now, we investigate how replacing BERTSUM with other newly proposed transformer-based summarization models affects the performance. More specifically, we use the following pre-trained models for fine-tuning to generate query focused abstractive summaries.

\textbf{BART:} BART \cite{bart} is a sequence-to-sequence model based on the transformer architecture. It was pre-trained based on denoising objectives to map a corrupted document to its original form. To pre-train BART for the original document reconstruction, the following objectives were utilized: \textit{document rotation, sentence permutation, text-infilling, token masking,} and \textit{token deletion}. We choose the pre-trained BART model since fine-tuning this model was found to be effective for the text generation task. Moreover, it utilizes a bidirectional encoder similar to the BERT encoder, while using a left-to-right autoregressive decoder.  

\textbf{PEGASUS:} This is another transformer-based encoder-decoder model that we choose for analysis because it is  particularly designed for abstractive summarization \cite{zhang2019pegasus}. For its pre-training objective, it resembles the downstream abstractive summarization task which involves generating summary-like text from an input document. To do so, it first selects and masks some sentences from the input document(s). Then it concatenates these selected sentences together to use them as a pseudo-summary. To select these sentences, the PEGASUS model investigates different approaches, such as: (i) randomly selecting $m$ sentences from the input document, (ii) selecting the first $m$ sentences in the input document, (iii) computing the ROUGE-1 score between each sentence and the rest of the document to select the top $m$ scored sentences. Using one of these approaches\footnote{It was empirically found that the approach that computed the ROUGE-1 score to select the top $m$ scored sentences was more effective \cite{zhang2019pegasus}.}, it identifies the sentences that are more important to the document and utilizes them as a pseudo reference summary for self-supervised learning. This way of self-supervised pre-training on large datasets leads to better and faster fine-tuning performance on various downstream abstractive summarization datasets. 

\textbf{T5:} The T5 model \cite{t5} is also a transformer model based on the BERT architecture. However, contrary to the traditional BERT-based models \cite{devlin2018bert,liu2019roberta,lan2019albert} that classify the given input text to a class label, the T5 model treats all tasks, such as, \textit{neural machine translation}, \textit{text classification}, \textit{question answering}, or \textit{text summarization} as a sequence to sequence problem. The model is pre-trained on a large dataset with different training and masking objectives to identify the best pre-training objective. The pre-trained model is then fine-tuned to generate the correct output for a given input sequence for the required task.

To use these models for our case study, we use the \textit{Hugging Face Transformer} \cite{Wolf2019HuggingFacesTS} for implementation. We compare the results of these new variations with our originally proposed BERTSUM-based  PreQFAS\textsubscript{WSL} model as well as the current state-of-the-arts in different evaluation metrics. \change{We show the results of our experiments in Table \ref{tab:csmdqfasf} to find that these new transformer-based models are also effective when used with our proposed PreQFAS\textsubscript{WSL} architecture. More importantly, some of these models even obtain new state-of-the-art results in different datasets. 

More specifically, we find that the {PreQFAS\textsubscript{WSL} model with T5} sets a new state-of-the-art in the DUC 2005 dataset in terms of both F1 and Recall in all ROUGE scores. In the DUC 2006 dataset, we find that the variation that uses the {PEGASUS model with {PreQFAS\textsubscript{WSL}}} sets a new state-of-the-art in terms of ROUGE-1 and ROUGE-2 metrics based on both Recall and F1. In terms of ROUGE-SU4, though our {PreQFAS\textsubscript{WSL} model with T5} sets a new state-of-the-art in the DUC 2006 dataset based on the F1 metric, it fails to outperform the current state-of-the-art RSA model \cite{baumel2018query} based on the Recall metric. In the DUC 2007 dataset, we again find that the {PreQFAS\textsubscript{WSL} model with T5} sets a new state-of-the-art in terms of all ROUGE scores based on the F1 metric. However, in terms of Recall, none of our models could outperform the current state-of-the-art models, \cite{dualces} and \cite{baumel2018query}, for both ROUGE-1 and ROUGE-SU4, respectively. Based on Recall for ROUGE-2, though both the T5 model and the PEGASUS model with PreQFAS\textsubscript{WSL} outperform all prior work, the {PreQFAS\textsubscript{WSL} model with T5} performs better than its counterpart { with PEGASUS} to set the new state-of-the-art result.

\seqvsbase 

From this case study, we find that fine-tuning different pre-trained transformers in the proposed PreQFAS\textsubscript{WSL} model is also useful for the QFAS task. This demonstrates the effectiveness of our proposed approach across various transformer-based summarization models. We conduct some additional case studies for the PreQFAS\textsubscript{WSL} model to investigate (i) \textit{how weak supervision by different pre-trained transformer models} and (ii) \textit{how ranking the summary sentences in the final stage by different answer selection models} may impact the overall performance. The results from these case studies can be found in Appendix B and Appendix C respectively.}

\paragraph{\textbf{(ii) Varying the Number of Gold Reference Summaries \textbf{\textit{K}} in PreQFAS\textsubscript{SFT}:}} In this case study, contrary to the previous analysis where we investigate the performance of the PreQFAS\textsubscript{WSL} model, we study the performance of the other variant of the PreQFAS architecture: the {PreQFAS\textsubscript{SFT}} model. To do so, we run different experiments with different number of gold reference summaries $k$. In other words, we vary the total number of fine-tuning runs in the PreQFAS\textsubscript{SFT} model where each fine-tuning run contains a different gold reference summary than other runs. Moreover, we use the following baseline for this case study where batch-wise\footnote{It refers to the traditional training procedure of neural network models where the input is given to the model in different batches.} fine-tuning is used: {BERTSUM\textsubscript{}}. For the batch-wise fine-tuning, instead of utilizing sequential fine-tuning by varying the gold reference summaries in different fine-tuning runs, we run the fine-tuning only once by using different gold reference summaries for the same input document in different batches. 

From Figure \ref{fig:SeqVsBase}, we find that in all datasets based on F1 and Recall, using multiple gold reference summaries improves the ROUGE-1 score in both PreQFAS\textsubscript{SFT} (i.e., sequential fine-tuning) and BERTSUM\textsubscript{} (i.e., batch-wise fine-tuning) models. However, the performance gain via increasing the number of gold reference summaries in our proposed PreQFAS\textsubscript{SFT} model is more than the baseline BERTSUM\textsubscript{}. More specifically, the maximum improvement from $k = 1$ to $k = 4$ in our proposed PreQFAS\textsubscript{SFT} model is 33.23\% (obtained in DUC 2007 in terms of the Recall metric) while in the baseline BERTSUM\textsubscript{} model is 5.71\% (obtained in DUC 2006 in terms of the F1 metric). Furthermore, in terms of F1, the average improvement from $k = 1$ to $k = 4$ in the PreQFAS\textsubscript{SFT} model is 14.44\% while in the BERTSUM\textsubscript{} model is 4.99\%. The improvement in terms of Recall is even more, as we find that the average improvement from $k = 1$ to $k = 4$ in the PreQFAS\textsubscript{SFT} model is 30.90\% while in the BERTSUM\textsubscript{} model is 7.18\%. This case study demonstrates the effectiveness of our proposed sequential fine-tuning technique with multiple gold reference summaries.

\csmdqfash 

\subsubsection{\textbf{Human Evaluation}} \label{section:human}

So far, we primarily use the ROUGE scores \cite{rouge} to evaluate our proposed models which are computed based on exact matches between the tokens of the generated summary and the gold reference summaries. As a consequence, ROUGE scores become lower when the generated summary contains tokens that are semantically similar to the gold reference summaries but not an exact match. Moreover, it fails to consider other important factors such as how much informative or fluent is the generated summary, as well as whether the generated summmary maintains coherence. Therefore, we also conduct human evaluation on Amazon Mechanical Turk\footnote{\url{https://www.mturk.com/}} for a qualitative analysis of our proposed models. For this purpose, we randomly selected 10 document sets from each of the three DUC datasets (2005-07). Thus, a total of 30 document sets was selected with each dataset containing 10 document sets. For each document set, we selected 3 human annotators who were asked to rate the summaries of different models with a score in between 1 to 5 (inclusive) based on the following three metrics:\\ \change{
\textit{\textbf{(i) Informativeness:} It measures how much informative is the generated summary.}\\ 
\textit{\textbf{(ii) Coherence:} In a coherent summary the sentences in the generated summary maintain consistent connection between them.}\\
\textit{\textbf{(iii) Fluency:} A fluent summary contains sentences that are grammatically correct.}}

For the human evaluation, we use the summaries generated by all variations of the PreQFAS\textsubscript{WSL} model: {BERTSUM}, {BART}, {PEGASUS}, and {T5}. In addition, we use the summaries generated by the PreQFAS\textsubscript{SFT} model that utilizes the BERTSUM model. We show the results of the human evaluation\footnote{The cases when at least two out of three annotators agreed on their ratings are 61\%, 44\%, and 60\% for coherence, fluency, and informativeness respectively.} in Table \ref{tab:csmdqfashuman}. We find from this table that even though the \textbf{PreQFAS\textsubscript{WSL} - {BART}} could not set a new state-of-the-art result in terms of different ROUGE scores (see Table \ref{tab:csmdqfasf}), it performs better than all other models based on all human evaluation metrics in DUC 2005 and DUC 2006 datasets. In DUC 2007, we find that in terms of Coherence, \textbf{PreQFAS\textsubscript{WSL} - {PEGASUS}} performs the best while in terms of Informativeness, the \textbf{PreQFAS\textsubscript{WSL} - {T5}} performs the best. In terms of Fluency in DUC 2007, we find that both \textbf{PreQFAS\textsubscript{WSL} - {T5}} and \textbf{PreQFAS\textsubscript{WSL} - {BART}} perform the best as they obtain the highest fluency score of 4.17. 

Furthermore, we observe that the PreQFAS\textsubscript{SFT} model performs poorer in most cases than all other models that are based on PreQFAS\textsubscript{WSL}. This may suggest that the utilization of weakly supervised learning to fine-tune the pre-trained model on each individual document provides more human readable summaries than its counterpart that applies filtering on the input document set for summary generation. Moreover, the superior performance of the PreQFAS\textsubscript{WSL} model that utilizes BART over other models gives a strong indication that the quality of the generated summaries of this model is better in terms of informativeness, fluency, and coherence.

\section{Conclusions and Future Work} \label{section:conclusion}

In this paper, we have presented a series of domain adaptation techniques from pre-trained transformer-based models to address the challenge of lack of training data for the query focused abstractive text summarization task. For the single-document scenario, we perform domain adaptation by pre-training a transformer-based model on a large dataset for generic abstractive summarization followed by fine-tuning it via incorporating the query relevance. For the multi-document scenario, we have presented two domain adaptation techniques that tackle the computational complexity problem for long text sequences. The first approach generates the weak reference summary of each document in the document set using distant supervision to fine-tune the pre-trained summarization model on each document to generate the query focused abstractive summary. The second approach filters out the sentences in the multi-document set to feed only the sentences that are most relevant to the query as input to the pre-trained transformer model. Then, we sequentially fine-tune the pre-trained transformer model using all gold reference summaries for a given multi-document set.   

We conducted extensive experiments with different variants of transformer-based models and different types of attention mechanisms for incorporating query relevance in the summarization models. Moreover, we conducted a series of ablation studies as well as case studies to carefully investigate the advantages of our proposed architecture. Additionally, we conducted human evaluations in all query focused multi-document summarization datasets to get a better understanding of the quality of the generated summaries of our proposed models.

To the best of our knowledge, the work presented in this paper is the first to give a comprehensive overview of how  domain adaptation from pre-trained transformers can be effectively utilized to tackle the few-shot learning problem in both single-document and multi-document query focused summarization datasets. Our experiments show that utilizing transfer learning from a transformer model pre-trained on a large dataset for generic abstractive summarization and then fine-tuning on the target query focused summarization dataset results in significant performance gain, setting new state-of-the-art results. In addition, our analysis reveals several new insights including the limitations of the Debatepedia dataset, the superior performance of recent transformer-based models such as T5 \cite{t5} and PEGASUS \cite{zhang2019pegasus} over their counterparts in the fine-tuning stage and the weakness of query-document attention mechanism compared to the bidirectional self-attention mechanism in some datasets. \change{We also analyze the memory requirements of our proposed architecture (see Appendix D for details) and find that it does not require huge computational resources for real-world production deployments.} 

Our findings in this paper lead to several new directions for future work. First, our work demonstrates the pressing need for constructing a new query focused single-document abstractive summarization dataset as we find that the existing benchmark dataset for the single-document query focused summarization task (i.e., Debatepedia dataset) is more of a generic summarization dataset and many queries in this dataset have no relation with the reference summaries. \change{Second, we will investigate the performance of our proposed approach with other domain adaptation
mechanisms such as adversarial losses and reweighting 
\cite{ramponi2020neuraldomainadaptationsurvey}. We also aim to investigate the performance of our approach  
in additional domains \cite{jin2020deep,zhong2021qmsum,vig2021exploring,bari2021uxla,sanh2021multitask}, and with other evaluation metrics \cite{louis2013evaluationsummaryautomatically,xenouleas2019sumqe,zhang2019bertscore,yuan2021bartscore}.} Third, we will study our proposed approach using other transformer models \cite{gpt2,tay2020efficient,zaheer2020bigbird,beltagy2020longformer,gpt3} and explore the performance in other related tasks, such as visual question answering \cite{VQA_antol}, chart question answering \cite{kim2020answering}, knowledge base question answering \cite{zhou2021dfm, asnq}, sentiment analysis \cite{JH5,JH6},
and biomedical information retrieval \cite{JH1,JH4}. Finally, we hope that  our source code that we will make publicly available will facilitate other researchers in reproducing our work and help to push the state-of-the-art in the future research of query focused abstractive summarization.

\section{Bibliographic Note}

Portions of this work have been published as short papers at the COLING 2020 \cite{laskar2020wslds} and the Canadian AI 2020 \cite{laskar2020query} conference proceedings. However, this work substantially extends the published papers in several ways, most notably:
(i) we investigate the performance of different attentions in query focused summary generation (Section \ref{section:preqfas_fsl}); (ii) we propose a novel sequential fine-tuning approach to utilize all the available multi-document gold reference summaries for supervised training (Section \ref{section:preqfas_cc}),  (iii) for the query focused abstractive summarization task in single-document scenarios, we conduct several ablation tests to investigate the effectiveness of different components used in our model (Section \ref{section:as_sdqfas}) as well as case studies to analyze the effectiveness of our model in zero-shot learning setup (Section \ref{section:cs_sdqfas}), as well as summarize the key limitations of the Debatepedia dataset (Section \ref{section:analyze_debatepedia}); (v) for the multi-document scenario, we study how incorporating recent transformer-based pre-trained summarizers in our proposed model impact the performance (Section \ref{section:cs_mdqfas}); and finally, (vi) in addition to extensive experiments on benchmark datasets, we also conduct human evaluation to qualitatively compare among different models proposed for query focused multi-document summarization (Section \ref{section:human}). Besides these extensions, the related work section was updated and a significant portion of this article was rewritten to adapt to a journal-style publication.

\section*{Acknowledgments}
We gratefully appreciate the associate editor and the reviewers of the \textit{Computational Linguistics} journal for their valuable and detailed comments that helped to improve the quality of this article. This work was done when the first author was a Research Assistant at the Information Retrieval and Knowledge Management Research Lab, York University, Canada. His research is supported by the Natural Sciences \& Engineering Research Council (NSERC) of Canada, the York Research Chairs (YRC) program and an ORF-RE (Ontario Research Fund-Research Excellence) award in BRAIN Alliance. We also acknowledge Compute Canada for providing us the computing resources to conduct experiments.

\change{

\section*{Appendix}

\appendix

\section{{Using different datasets for pre-training (Case Study)}} 
Here, we investigate whether using different datasets to pre-train the proposed PreQFAS model can lead to different results. For that purpose, we use two pre-training datasets: (i) XSUM, and (ii) CNNDM. After pre-training two BERTSUM models in these two datasets, we fine-tune both pre-trained models on the MS-MARCO dataset (the fine-tuned models are then used to generate the summaries in the MediQA-AnS datasets), as well as fine-tune these models on the Debatepedia dataset (the fine-tuned models are then used to generate the summaries in the target Debatepedia dataset).

\zssdqfasG

Table \ref{tab:zsqfasG} shows that in all evaluation datasets, using the CNN-DM dataset for pre-training is always better than using the XSUM dataset in terms of all ROUGE scores. This may indicate that using a larger-sized dataset for pre-training is more helpful (the CNN-DM dataset has 2,87,227 training instances whereas the XSUM dataset contains 2,04,045 training instances). }

\change{\section{Utilizing different models for Weak Supervision (Case Study)}}


As mentioned earlier (see Section \ref{section:wsl_ds}), in our PreQFAS\textsubscript{WSL} model, we generate the initial weak reference summaries using the RoBERTa model \cite{liu2019roberta} fine-tuned on the MS-MARCO dataset \cite{msmarco} for the answer selection task \cite{laskar-LREC}. To examine the effectiveness of utilizing other models for weak supervision, we use the following variations:

   \textbf{BERTSUM\textsubscript{EXT}}: For this variation, we adopt the pre-trained BERTSUM  model from \cite{liuemnlpbertsum} that was trained for the generic extractive summarization task in the CNN-DM dataset \cite{cnndm}. For each dataset, the initial weak reference summaries are first generated using this model. Then, distant supervision from the multi-document gold reference summaries is applied using the RoBERTa\textsubscript{MRPC} model to generate the weak abstractive reference summary of each training document. 
 
 \textbf{BERTSUM\textsubscript{ABS-EXT}}: 
    For this variation, we adopt the pre-trained BERTSUM model from \citet{liuemnlpbertsum} that was trained for the generic abstractive summarization task (after being initially trained for extractive summarization) in the CNN-DM dataset \cite{cnndm}. Then, for each dataset, the initial weak reference summaries are generated using this model. Afterward, we applied distant supervision from the multi-document gold reference summaries using the RoBERTa\textsubscript{MRPC} model to generate the weak abstractive reference summary of each training document.

\csmdqfasfwH

We show the result of our experiments in Table \ref{tab:csmdqfaswH} and find that for ROUGE-1 and ROUGE-SU4 (in terms of both F1 and Recall), the new variants that utilized the BERTSUM\textsubscript{ABS-EXT} and the BERTSUM\textsubscript{EXT} models for initial weak reference summary generation could not outperform the original architecture that utilized RoBERTa. Based on ROUGE-2, we find that in both DUC 2005 and 2007 datasets, the RoBERTa model  outperforms the BERTSUM-based weak reference summary generators (for both F1 and Recall). The only scenario when RoBERTa does not perform the best in this metric is in the DUC 2006 dataset, where the BERTSUM\textsubscript{EXT} model outperforms other variants in terms of both F1 and Recall metrics. Though the RoBERTa model performs better than other variations in most scenarios for the initial weak reference summary generation (based on different datasets and evaluation metrics), the difference between this model with other variants is \textbf{not statistically significant} based on paired t-test ($p$ $>$ $.05$).\\

\change{\section{Using different datasets for fine-tuning the Answer Selection model (Case Study)}}
During the final stage of the  PreQFAS\textsubscript{WSL} model, we select the most relevant sentences as the query focused abstractive summary using the RoBERTa\textsubscript{Large} model fine-tuned for the answer selection task. While we originally choose to fine-tune the pre-trained RoBERTa\textsubscript{Large} model on the MS-MARCO dataset for answer selection, here we examine how varying the dataset for fine-tuning the RoBERTa\textsubscript{Large} model affects the performance. For this purpose, we utilize the following question answering datasets.

\csmdqfasansI

\textbf{TREC-QA:} This dataset is created from the QA track (8-13) of Text REtrieval Conference~\cite{wang2007jeopardy}. It contains 1,229 questions with 53,417 candidate answers.

\textbf{WikiQA:} This is an open domain QA
dataset \cite{yang2015wikiqa} in which the answers
were collected from the Wikipedia. The WikiQA dataset contains 2,118 questions with 20,360 candidate answers.

\textbf{SemEvalCQA (2015):} This is a Community Question Answering (CQA) dataset that has been created from \textit{Qatar Living Forums\footnote {\url{https://www.qatarliving.com/forum}}}. It contains 2,600 questions and 16,541 candidate answers in the training set. 


\textbf{SemEvalCQA (2016 \& 2017):} The SemEvalCQA-2016 and the SemEvalCQA-2017 datasets are also CQA datasets created from \textit{Qatar Living Forums}. Both datasets have the same training data containing 4,879 questions and 36,198 candidate answers.

We show the results of our experiments in Table \ref{tab:csmdqfasansI} to find that in terms of both Recall and F1, the original model that was fine-tuned on the MS-MARCO dataset for the final summary selection performs better than the variations that were find-tuned on other datasets. This could be because the MS-MARCO dataset consists of 1,53,725 queries and 15,37,250 candidate answers that are much larger than other datasets.\\

\change{\section{Model requirements} 

In this work, we keep our models reasonably light-weight  by demonstrating the effectiveness of our proposed approach via utilizing a simple transformer-based summarization architecture, the BERTSUM model which utilizes the BERT model as encoder and the decoder of Transformer as decoder. The size of the trained model is only 2.32GB that makes it feasible for the single-document query focused abstractive summarization task in industrial production scenarios in a computationally limited resource environment. 

For our proposed approaches for multi-document scenarios, we additionally use a RoBERTa model that only contains 1.43GB of additional memory. In total, our proposed system can be used for production use-cases in a very lightweight machine that does not require more than 4GB of RAM to host our models (1.43GB RoBERTa model for sentence filtering or final summary selection, while 2.32GB BERTSUM model for summarization). In more powerful computing environments, other models that provide superior performance (e.g., BART, PEGASUS, T5 etc.) can be used too.}

\change{
\section{The Rationale behind generating the initial weak extractive summaries} 

To generate the weak abstractive reference summary, we apply the RoBERTa sentence similarity model to measure the similarity between each sentence in the weak extractive reference summary and each sentence in the gold reference summaries. However, instead of generating the weak extractive reference summary, if we compare the similarity between each sentence in the multi-document set with each sentence in the gold reference summaries, the model will take a considerable amount of time. In our experiments for multi-document scenarios, each DUC dataset consists of up to 37,925 sentences on average, whereas the average number of sentences in the gold reference summaries per dataset may contain only about 3,089 sentences. Thus, if there are $M$ sentences in the document set and $N$ sentences in the gold reference summaries, the total time complexity of the similarity modeling will be $O(M*N)$. This will make the weak reference summary generation process very time consuming that may not be acceptable in real-world industrial scenarios where models are required to be trained regularly. 

Therefore, to avoid this huge computation, for each document in a document set we first generate the initial weak reference summary (3 sentences long) using a pre-trained model. And then we measure the similarity between each sentence in the generated weak extractive reference summary with each sentence in the gold reference summaries. This step saves a huge computational time for weak abstractive reference summary generation for the training stage. }

\bibliography{references}

\bibliographystyle{acl_natbib}

\end{document}

%% file: tables.tex
\newcommand{\introductiontable}{
\begin{table*}

\caption[An example of the Query Focused Text Summarization task]{
An example of the Query Focused Text Summarization Task to generate the abstractive summary from the given source document.}

\centering

\begin{tabular}{|p{12cm}|}
\hline
\textbf{Query:}  What is the benefit of reality shows?\\
\textbf{Document:} Even if reality shows were not enlightening, they generate massive revenues that can be used for funding more sophisticated programs. Take BBC for example, it offers entertaining reality shows such as total wipeout as well as brilliant documentaries.\\
\textbf{Summary:} Reality show generates revenues.\\
\hline
\end{tabular}

\label{tab:introduction} 
\end{table*}}

\newcommand{\canaitable}{
\begin{table*}

\caption[Performance comparisons on the Debatepedia dataset]{Performance of different models for the SD-QFAS task on the Debatepedia dataset. Here, `R', `P', and `F' denote `Recall', `Precision', and `F1' respectively, while `QD' \space denotes `Query-Document Attention' and `BSA' denotes \space `Bidirectional Self-Attention'. The results for the DDA, the Selection Driven, the Overlap-Wind, and the RSA model are collected from \cite{nema}, 
\cite{canaiqfas}, \cite{ecirqfas}, and \cite{baumel2018query} respectively.} 

\centering

\begin{tabular}
{|c|c|c|c|c|c|c|c|c|c|}
\hline
\multicolumn{1}{|c|}{\textbf{MODEL}} &
\multicolumn{3}{c|}{\textbf{ROUGE-1}} & 
\multicolumn{3}{c|}{\textbf{ROUGE-2}} & 
\multicolumn{3}{c|}{\textbf{ROUGE-L}} \\ \cline{2-10}
\centering
 & \textbf{R} & \textbf{P} & \textbf{F} 
& \textbf{R} & \textbf{P} & \textbf{F} 
& \textbf{R} & \textbf{P} & \textbf{F} 
\\ \hline

\textbf{QR-BERTSUM\textsubscript{Vanilla}} & 22.3 & 35.7 & 26.4 & 9.9 & 16.7 & 11.9 & 21.2 & 33.9 & 25.1
\\ \hline

\textbf{DDA 
} & 41.3 & - & - & 18.8 & - & - & 40.4 & - & - 
\\ \hline

\textbf{Selection Driven 
} & 43.2 & - & - & 27.4 & - & - & 42.7 & - & - 
\\ \hline

\textbf{Overlap-Wind 
} & 44.4 & - & - & 30.5 & - & - & 44.2 & - & - 
\\ \hline

\textbf{RSA 
} & 53.1 & - & - & 16.1 & - & - & 46.2 & - & - 
\\ \hline

\textbf{PreQFAS (QD)} & \textbf{58.0} & 60.3 & \textbf{58.7} & \textbf{45.2} & \textbf{46.1} & \textbf{45.5} & \textbf{57.1} & 59.2 & \textbf{57.7}
\\\hline

\textbf{PreQFAS (BSA)} & \textbf{58.0} & \textbf{60.4} & 58.5 & \textbf{45.2} & \textbf{46.1} & \textbf{45.5} & \textbf{57.1} & \textbf{59.3} & \textbf{57.7}
\\\hline

\end{tabular}
\label{tab1} 

\end{table*}}

\newcommand{\zssdqfas}{
\begin{table*}

\caption{\change{Case study results to investigate the zero-shot learning performance based on the F1 metric. In this experiment, the BERTSUM\textsubscript{XSUM} is used as a baseline model that was pre-trained only on the XSUM dataset without any fine-tuning while the PreQFAS\textsubscript{MS-MARCO} model was first pre-trained on XSUM and then fine-tuned on MS-MARCO. Here, we denote `ROUGE' as 'R'.}}
\centering

\begin{tabular}
{|c|c|c|c|c|c|c|c|c|c|}
\hline
& \multicolumn{9}{|c|}{\textbf{Datasets}} \\ \cline{2-10}

\multicolumn{1}{|c|}{\textbf{MODEL}} &
\multicolumn{3}{c|}{\textbf{MediQA-AnS (Pages)}} & 
\multicolumn{3}{c|}{\textbf{MediQA-AnS (Passages)}} & 
\multicolumn{3}{c|}{\textbf{Debatepedia}} \\ \cline{2-10}
\centering
 & \textbf{R-1} & \textbf{R-2} & \textbf{R-L} 
& \textbf{R-1} & \textbf{R-2} & \textbf{R-L} 
& \textbf{R-1} & \textbf{R-2} & \textbf{R-L} 
\\ \hline

\textbf{BERTSUM\textsubscript{XSUM}} & 19.87 & 3.59 & 13.49 & 21.36 & 4.02 & 13.95 & 13.3 & 2.8 & 11.5
\\ \hline

\textbf{PreQFAS\textsubscript{MS-MARCO}} & \textbf{23.07} & \textbf{5.41} & \textbf{15.35} & \textbf{29.89} & \textbf{11.29} & \textbf{21.05} & \textbf{22.2} & \textbf{6.0} & \textbf{19.7}

\\\hline

\end{tabular}
\label{tab:zsqfas} 

\end{table*}}


\newcommand{\msmarcoqfastable}{
\begin{table*}

\caption[Performance comparisons on the MS-MARCO dataset]{Performance of different models for the SD-QFAS task on the MS-MARCO dataset in terms of ROUGE-L and BLEU-1 based on the F1 metric. }

\centering

\begin{tabular}{|c|c|c|}
\hline

\textbf{MODEL} & \textbf{ROUGE-L} & \textbf{BLEU-1}
\\ \hline

\textbf{QR-BERTSUM\textsubscript{Vanilla}} & 71.6 & 70.2 
\\ \hline

\textbf{MASQUE \cite{nishida2019multi}} & \textbf{78.7} & 78.1 
\\ \hline

\textbf{PreQFAS (QD Attention)} & 72.3 & 72.1
\\\hline

\textbf{PreQFAS (Bidirectional Self-Attention)} & 78.4 & \textbf{80.4} 
\\\hline

\end{tabular}

\label{tab2} 

\end{table*}}



\newcommand{\ablationqfas}{
\begin{table*}

\centering

\caption[Ablation test on Debatepedia and MS-MARCO datasets]{Ablation test results in terms of Recall on Debatepedia and in terms of F1 on MS-MARCO. For this ablation test, all models were pre-trained on the XSUM dataset. Since removing the query relevance as well as fine-tuning makes both PreQFAS models (based on attention) the same, we only mention the result once for the Bidirectional Self-Attention. Here, we denote `Bidirectional Self-Attention' as `BSA' and `QD attention' as `QD', while `w/o' denotes `without'. }

\label{tab:ablationqfas}. 


\begin{tabular}{|l|c|c|c|c|c|}

\hline
\multicolumn{1}{|l|}{} &
\multicolumn{5}{c|}{\textbf{Datasets}} \\ \cline{2-6}
\multicolumn{1}{|c|}{{\textbf{MODEL}}} &
\multicolumn{3}{c|}{\textbf{Debatepedia}} & \multicolumn{2}{c|}{\textbf{MS-MARCO}} \\ \cline{2-6}
\centering
 & \textbf{\textbf{ROUGE-1}} & \textbf{ROUGE-2} & \textbf{\textbf{ROUGE-L}} & \textbf{BLEU-1} & \textbf{\textbf{ROUGE-L}}

\\ \hline

\textbf{PreQFAS (BSA)} & 57.96 & 45.20 & 57.05 & 80.39 & 78.39 
\\

\textbf{PreQFAS (QD)} & 57.97 & 45.21 & 57.06 & 72.10 & 72.25 
\\ \hline

\textbf{w/o Query Relevance}  & 56.82 & 44.66  & 56.07  & 66.21 &  61.50
\\ \hline

\textbf{w/o Fine-Tuning} & 17.36 & 11.48 & 13.32 & 20.14 & 21.52
\\
\hline

\end{tabular}

\end{table*}}



\newcommand{\daqfas}{
\begin{table*}

\centering

\caption{Zero-shot learning performance on the Debatepedia dataset. Here, `FT' denotes `Fine-Tuning'.}\label{tab:daqfas} 


\begin{tabular}{|c|c|c|c|c|c|c|}

\hline
\multicolumn{1}{|c|}{} &
\multicolumn{6}{c|}{\textbf{Pre-training Datasets}} \\ \cline{2-7}
\multicolumn{1}{|c|}{{\textbf{MODEL}}} &
\multicolumn{3}{c|}{\textbf{CNN-DM}} & \multicolumn{3}{c|}{\textbf{XSUM}} \\ \cline{2-7}
\centering
 & \textbf{\textbf{R-1}} & \textbf{R-2} & \textbf{\textbf{R-L}} &  \textbf{\textbf{R-1}} & \textbf{R-2} & \textbf{\textbf{R-L}}

\\ \hline
\textbf{PreQFAS\textsubscript{Debatepedia} FT} & 57.96 & 45.20 & 57.05 & 80.39 & 78.39 & 1.2
\\

\textbf{PreQFAS\textsubscript{MS-MARCO} FT} & 30.49 & 9.18 & 25.97 & 80.39 & 78.39 & 1.2
\\

\hline

\end{tabular}


\end{table*}}


\newcommand{\analyzedebatepedia}{
\begin{table*}
\centering
\caption{Some examples from the Debatepedia dataset.}
\begin{tabular}{|p{15cm}|}
\hline
\textbf{(a) Query having no relevance with the document or the summary} \\
\hline
\textbf{Query:} Does an MBA enhance leadership skills?\\
\textbf{Document:} Business schools might improve your quantitative presentation and communication skills. It might but get you thinking about ethical and strategy. But two years of case studies aren't go to turn you into a leader if you weren't died one. There's no learning charisma persuasiveness elegance or gut instinct.\\
\textbf{Gold Summary:} PhD will not improve cm factors of leaders. \\
\hline
\textbf{(b) Query is a Yes/No type close-ended question} \\
\hline
\textbf{Query:} Is investing in new technologies desirable?\\
\textbf{Document:} Student will neglect their thought skill and rely too much on technology for everything.\\
\textbf{Gold Summary:} Spending cash on technologies is a waste. \\
\hline
\textbf{(c) Query is relevant to the document but the summary is quite generic} \\
\hline
\textbf{Query:} Is merit-based pay fair? \\
\textbf{Document:} Merit pay creates an incentive for teachers to cheat by improving student test scores so that they can appear to be doing better as a result of the teacher's work resulting in bonuses and higher pay. Obviously, the resulting differences in pay would not be fair. \\
\textbf{Gold Summary:} Merit pay motivates teachers to cheat on test-scoring. \\
\hline
\end{tabular}

\label{tab:analyzedebatepedia} 
\end{table*}}

\newcommand{\mdqfasfive}{
\begin{table*}

\caption[Performance comparisons on the DUC 2005 dataset]{Performance comparisons in terms of (a) \textbf{F1} and (b) \textbf{Recall} on the DUC 2005 dataset. Here, `*' denotes extractive summarization model, while `$\dagger$', `$\mp$', and `$\S$' indicate that the initial weak reference summaries are generated by BERTSUM\textsubscript{EXT}, BERTSUM\textsubscript{ABS-EXT}, and RoBERTa\textsubscript{MS-MARCO} respectively. Here, we denote `ROUGE' as 'R'.}

\centering
\begin{tabular}{|c|c|c|c|c|c|c|}
\hline

\multicolumn{1}{|c|}{} &
\multicolumn{6}{c|}{\textbf{DUC 2005}}  \\ \cline{2-7}

\multicolumn{1}{|c|}{\large{\textbf{Model}}} &
\multicolumn{3}{c|}{\textbf{F1}} & 
\multicolumn{3}{c|}{\textbf{Recall}} \\ \cline{2-7}

 & \textbf{\textbf{R-1}} & \textbf{R-2} & \textbf{\textbf{R-SU4}} 
& \textbf{\textbf{R-1}} & \textbf{R-2} & \textbf{\textbf{R-SU4}}

\\ \hline

\textbf{CES-50 \cite{queryfocusedsummarization2017unsupervised} *} & 37.78 & 7.45 & 13.02 & 40.35 & 7.94 & 13.91 
\\ 

\textbf{RSA \cite{baumel2018query} } & - & - & - & 39.82 & 6.98 & \textbf{15.73}
\\ 

\textbf{Dual-CES \cite{dualces} *} & 38.08 & 7.54 & 13.17 & \textbf{40.82} & 8.07 & 14.13 
\\

\hline

\textbf{BERTSUM\textsubscript{EXT} *} & 37.52 & 7.84 & 13.29 & 37.55 & 7.84 & 13.31 
\\ 

\textbf{BERTSUM\textsubscript{ABS}} & 38.35 & 7.94 & 13.44 & 38.36 & 7.92 & 13.43 
\\

\hline

\textbf{PreQFAS\textsubscript{WSL} $\dagger$} & 40.13 & 8.94 & 14.53 & 40.16 & 8.94 & 14.54 
\\ 

\textbf{PreQFAS\textsubscript{SFT} $\mp$} & 42.48 & 10.55 & 16.02 & 41.96 & 10.39 & 15.79 
\\

\hline
\end{tabular}
\label{tab:mdqfasfive} 
\end{table*}

}

\newcommand{\mdqfassix}{

\begin{table*}

\caption[Performance comparisons on the DUC 2006 dataset]{Performance comparisons in terms of (a) \textbf{F1} and (b) \textbf{Recall} on the DUC 2006 dataset. Here, `*' denotes extractive summarization model, while `$\dagger$', `$\mp$', and `$\S$' indicate that the initial weak reference summaries are generated by BERTSUM\textsubscript{EXT}, BERTSUM\textsubscript{ABS-EXT}, and RoBERTa\textsubscript{MS-MARCO} respectively. Here, we denote `ROUGE' as 'R'.}

\centering
\begin{tabular}{|c|c|c|c|c|c|c|}
\hline

\multicolumn{1}{|c|}{} &
\multicolumn{6}{c|}{\textbf{DUC 2006}}  \\ \cline{2-7}

\multicolumn{1}{|c|}{\large{\textbf{Model}}} &
\multicolumn{3}{c|}{\textbf{F1}} & 
\multicolumn{3}{c|}{\textbf{Recall}} \\ \cline{2-7}

 & \textbf{\textbf{R-1}} & \textbf{R-2} & \textbf{\textbf{R-SU4}} 
& \textbf{\textbf{R-1}} & \textbf{R-2} & \textbf{\textbf{R-SU4}}

\\ \hline

\textbf{CES-50 \cite{queryfocusedsummarization2017unsupervised} *} & 40.47 & 9.13 & 14.73 & 43.01 & 9.69 & 15.65
\\ 

\textbf{RSA \cite{baumel2018query} } & - & - & - & 42.89 & 8.73 & \textbf{17.75} 
\\ 

\textbf{Dual-CES \cite{dualces} *} & 41.23 & 9.47 & 14.97 & \textbf{43.94} & 10.09 & 15.96 
\\ 

\textbf{QUERYSUM \cite{xulapata2020coarseemnlp} *} & 41.6 & 9.5 & 15.3 & - & - & - 
\\

\hline

\textbf{BERTSUM\textsubscript{EXT} *} & 40.68 & 9.29 & 14.66 & 40.41 & 9.22 & 14.56 
\\ 

\textbf{BERTSUM\textsubscript{ABS}} & 40.87 & 9.43 & 14.83 & 40.59 & 9.39 & 14.73 
\\

\hline

\textbf{PreQFAS\textsubscript{WSL} $\dagger$} & 43.44 & \textbf{10.94} & \textbf{16.46} & 43.11 & \textbf{10.85} & 16.34 
\\ 

\textbf{PreQFAS\textsubscript{SFT} $\mp$} & 42.48 & 10.55 & 16.02 & 41.96 & 10.39 & 15.79 
\\

\hline
\end{tabular}
\label{tab:mdqfassix} 
\end{table*}

}

\newcommand{\mdqfasseven}{

\begin{table*}

\caption[Performance comparisons on the DUC 2007 dataset]{Performance comparisons in terms of (a) \textbf{F1} and (b) \textbf{Recall} on the DUC 2007 dataset. Here, `*' denotes extractive summarization model, while `$\dagger$', `$\mp$', and `$\S$' indicate that the initial weak reference summaries are generated by BERTSUM\textsubscript{EXT}, BERTSUM\textsubscript{ABS-EXT}, and RoBERTa\textsubscript{MS-MARCO} respectively. Here, we denote `ROUGE' as 'R'.}

\centering
\begin{tabular}{|c|c|c|c|c|c|c|}
\hline
\multicolumn{1}{|c|}{} &
\multicolumn{6}{c|}{\textbf{DUC 2007}}  \\ \cline{2-7}
\multicolumn{1}{|c|}{\large{\textbf{Model}}} &
\multicolumn{3}{c|}{\textbf{F1}} & 
\multicolumn{3}{c|}{\textbf{Recall}} \\ \cline{2-7}
& \textbf{\textbf{R-1}} & \textbf{R-2} & \textbf{\textbf{R-SU4}} 
& \textbf{\textbf{R-1}} & \textbf{R-2} & \textbf{\textbf{R-SU4}} 
\\ \hline

\textbf{CES-50 \cite{queryfocusedsummarization2017unsupervised} *} & 42.86 & 11.34 & 16.53 & 45.45 & 12.02 & 17.54 
\\ 

\textbf{RSA \cite{baumel2018query} } & - & - & - & 43.92 & 10.13 & \textbf{18.54} 
\\ 

\textbf{Dual-CES \cite{dualces} *} & 43.24 & 11.78 & 16.83 & \textbf{46.02} & \textbf{12.53} & 17.91 
\\ 

\textbf{QUERYSUM \cite{xulapata2020coarseemnlp} *} & 43.3 & 11.6 & 16.8 & - & - & - 
\\

\hline

\textbf{PreQFAS\textsubscript{EXT} *} & 42.57 & 11.20 & 15.98 & 42.41 & 11.08 & 15.92 
\\ 

\textbf{PreQFAS\textsubscript{ABS}} & 42.17 & 10.82 & 15.98 & 42.05 & 10.79 & 15.91 
\\

\hline

\textbf{PreQFAS\textsubscript{WSL} $\dagger$} & 44.29 & 11.89 & 17.24 & 44.11 & 11.84 & 17.16 
\\ 

\textbf{PreQFAS\textsubscript{WSL} $\mp$} & 44.20 & 11.80 & 17.12 & 43.72 & 11.53 & 16.92 
\\

\textbf{PreQFAS\textsubscript{WSL} $\S$} & \textbf{44.72} & \textbf{12.44} & \textbf{17.72} & 44.61 & 12.40 & 17.66 \\

\hline
\end{tabular}
\label{tab:mdqfasseven} 
\end{table*}
}

\newcommand{\mdqfasablationall}{
\begin{table*}

\caption{\change{Ablation Test result in terms of ROUGE-1 based on the average across all three datasets. Here, `without' is denoted by `w/o'.}}

\centering


\begin{tabular}{|l|l|l|l|}
    \hline
    \textbf{Model}&\textbf{Recall}&\textbf{F1}&\textbf{Statistically Significant}\\
  \hline
    \textbf{PreQFAS\textsubscript{WSL}} & \textbf{42.73} & \textbf{42.84} &  \\
        \textbf{w/o Distant Supervision} & 41.77 \tiny{(- 2.25\%)} & 41.88 \tiny{(- 2.24\%)} & {No (paired t-test, $p$ $>$ $.05$)} \\
        \textbf{w/o Trigram Blocking} & 40.92 \tiny{(- 4.24\%)} & 41.01 \tiny{(- 4.27\%)} & {No (paired t-test, $p$ $>$ $.05$)}  \\
         \textbf{w/o Weakly Supervised Learning} & 40.01 \tiny{(- 6.37\%)} & 40.12 \tiny{(- 6.35\%)} & \textbf{Yes} (paired t-test, $p$ $\leq$ $.05$)
         \\
          \hline
    \textbf{PreQFAS\textsubscript{SFT}} & {40.80} & {42.14} &  \\
        \textbf{w/o Fine-Tuning Sequentially} & {32.87 \tiny{(- 19.44\%)}} & {38.63 \tiny{(- 8.33\%)}} & \textbf{Yes} (paired t-test, $p$ $\leq$ $.05$) \\

 \hline

  \end{tabular}
\label{tab:mdqfasablationall} 

\end{table*}}

\newcommand{\wstable}{
\begin{table*}
\caption{\change{Performance comparisons in terms of \textbf{F1} and \textbf{Recall}. Here, `*' denotes extractive model. Moreover, the results for CES-50, QUERYSUM, DUAL-CES, and RSA are taken from \cite{queryfocusedsummarization2017unsupervised}, \cite{xulapata2020coarseemnlp}, \cite{dualces}, and \cite{baumel2018query} respectively. Here, we denote `ROUGE' as 'R'.}}

\begin{tabular}{|c|c|c|c|c|c|c|c|c|c|}
\cline{1-10}
 \multicolumn{10}{|c|}{\textbf{Metric: F1 Score}} \\
\cline{1-10}
\multicolumn{1}{|c|}{} & \multicolumn{9}{|c|}{\textbf{Datasets}} \\
\cline{2-10}
\multicolumn{1}{|c|}{\textbf{MODEL}} &
\multicolumn{3}{c|}{\textbf{DUC 2005}} & 
\multicolumn{3}{c|}{\textbf{DUC 2006}} & 
\multicolumn{3}{c|}{\textbf{DUC 2007}} \\ \cline{2-10}
\centering
\textbf{} & \textbf{\textbf{R-1}} & \textbf{R-2} & \textbf{\textbf{R-SU4}} 
& \textbf{\textbf{R-1}} & \textbf{R-2} & \textbf{\textbf{R-SU4}} 
& \textbf{\textbf{R-1}} & \textbf{R-2} & \textbf{\textbf{R-SU4}} 
\\ \hline
\textbf{CES-50 *} & 37.78 & 7.45 & 13.02 & 40.47 & 9.13 & 14.73 & 42.86 & 11.34 & 16.53 
\\ 
\textbf{QUERYSUM  *} & - & - & - & 41.6 & 9.5 & 15.3 & 43.3 & 11.6 & 16.8
\\ 
\textbf{DUAL-CES *} & 38.08 & 7.54 & 13.17 & 41.23 & 9.47 & 14.97 & 43.24 & 11.78 & 16.83
\\ 
\hline
\textbf{BERTSUM\textsubscript{EXT} *} & 37.52 & 7.84 & 13.29 & 40.68 & 9.29 & 14.66 & 42.57 & 11.20 & 15.98
\\ 
\textbf{BERTSUM\textsubscript{ABS}} & 38.35 & 7.94 & 13.44 & 40.87 & 9.43 & 14.83 & 42.17 & 10.82 & 15.98
\\
\hline
\textbf{PreQFAS\textsubscript{WSL}} & 40.32 & 9.17 & 14.71& \textbf{43.49} & \textbf{10.78} & \textbf{16.45} & \textbf{44.72} & \textbf{12.44} & \textbf{17.72} 
\\ 

\textbf{PreQFAS\textsubscript{SFT}} & \textbf{40.69} & \textbf{9.27} & \textbf{15.12} & 43.01 & 10.51 & 16.40 & 42.71 & 10.87 & 16.45
\\ 
\cline{1-10}
 \multicolumn{10}{|c|}{\textbf{Metric: Recall}} \\
\cline{1-10}
\multicolumn{1}{|c|}{} & \multicolumn{9}{|c|}{\textbf{Datasets}} \\
\cline{2-10}
\multicolumn{1}{|c|}{\textbf{MODEL}} &
\multicolumn{3}{c|}{\textbf{DUC 2005}} & 
\multicolumn{3}{c|}{\textbf{DUC 2006}} & 
\multicolumn{3}{c|}{\textbf{DUC 2007}} \\ \cline{2-10}

\textbf{} & \textbf{\textbf{R-1}} & \textbf{R-2} & \textbf{\textbf{R-SU4}} 
& \textbf{\textbf{R-1}} & \textbf{R-2} & \textbf{\textbf{R-SU4}} 
& \textbf{\textbf{R-1}} & \textbf{R-2} & \textbf{\textbf{R-SU4}} 
\\ \hline
\textbf{CES-50 *} & 40.35 & 7.94 & 13.91 & 43.01 & 9.69 & 15.65 & 45.45 & 12.02 & 17.54
\\ 
\textbf{RSA} & 39.82 & 6.98 & \textbf{15.73} & 42.89 & 8.73 & \textbf{17.75} & 43.92 & 10.13 & \textbf{18.54}
\\ 
\textbf{DUAL-CES  *} & \textbf{40.82} & 8.07 & 14.13 & \textbf{43.94} & 10.09 & 15.96 & \textbf{46.02} & \textbf{12.53} & 17.91
\\ 
\hline
\textbf{BERTSUM\textsubscript{EXT} *} & 37.55 & 7.84  &  13.31  &  40.41  &  9.22  & 14.56 &  42.41 &  11.08 &  15.92
\\ 
\textbf{BERTSUM\textsubscript{ABS}} & 38.36 & 7.92 & 13.43 & 40.59 & 9.39 & 14.73 & 42.05 & 10.79 & 15.91 
\\ 
\hline
\textbf{PreQFAS\textsubscript{WSL}} & 40.36 & \textbf{9.17} & 14.74 & 43.22 & \textbf{10.70} & 16.35 & 44.61 & 12.40 & 17.66
\\ 

\textbf{PreQFAS\textsubscript{SFT}} & 39.61 & 9.01 & 14.71 & 41.47 & 10.08 & 15.77 & 41.33 & 10.52 & 15.92
\\ 
\hline
\end{tabular}

  \label{tab:wstable} 

\end{table*}}

\newcommand{\csmdqfasf}{
\begin{table*}

\caption{\change{Case study results for the PreQFAS\textsubscript{WSL-DS} architecture in terms of F1 and Recall on the MD-QFAS datasets based on fine-tuning different models for summary generation. Here, we denote `ROUGE' as 'R'. For each dataset, the State-Of-The-Art (SOTA) result is taken from the following: for DUC 2005, all results are taken from \cite{dualces}; for DUC 2006, all results are taken from \cite{xulapata2020coarseemnlp}; for DUC 2007, R-1 is taken from \cite{xulapata2020coarseemnlp} while R-2 and R-SU4 are taken from \cite{dualces}.}} 
\centering

\begin{tabular}
{|c|c|c|c|c|c|c|c|c|c|}

 \hline
\multicolumn{10}{|c|}{\textbf{Metric: F1 Score}}  \\ \hline
& \multicolumn{9}{|c|}{\textbf{Datasets}} \\ \cline{2-10}

\multicolumn{1}{|c|}{\textbf{PreQFAS\textsubscript{WSL} model}} &
\multicolumn{3}{c|}{\textbf{DUC 2005}} & 
\multicolumn{3}{c|}{\textbf{DUC 2006}} & 
\multicolumn{3}{c|}{\textbf{DUC 2007}} \\ \cline{2-10}
\centering
 & \textbf{R-1} & \textbf{R-2} & \textbf{R-SU4} 
 & \textbf{R-1} & \textbf{R-2} & \textbf{R-SU4} 
 & \textbf{R-1} & \textbf{R-2} & \textbf{R-SU4} 
\\ \hline



\textbf{{\textit{with} BERTSUM}} & {40.3} & {9.3} & {14.7} & {43.5} & {10.8} & {16.5} & {44.7} & {12.4} & {17.7} 
\\\hline

\textbf{{\textit{with} BART}} & {40.3} & {8.7} & {14.8} & {43.1} & {10.2} & {16.1} & {44.3} & {11.4} & {17.0} 
\\\hline
\textbf{{\textit{with} PEGASUS}}  & {39.8} & {9.2} & {14.7} & \textbf{44.3} & \textbf{11.5} & {16.9} & {44.8} & {12.7} & {17.8} 
\\\hline
\textbf{{\textit{with} T5}} & \textbf{41.3} & \textbf{9.9} & \textbf{15.8} & {44.0} & {11.2} & \textbf{17.0} & \textbf{45.4} & \textbf{13.0} & \textbf{18.3} 
\\\hline
\textbf{SOTA} & 38.1 & 7.5 & 13.2 & 41.2 & 9.5 & 15.3 & 43.3 & 11.8 & 16.8

\\\hline

\hline
\multicolumn{10}{|c|}{\textbf{Metric: Recall}}  \\ \hline
& \multicolumn{9}{|c|}{\textbf{Datasets}} \\ \cline{2-10}

\multicolumn{1}{|c|}{\textbf{PreQFAS\textsubscript{WSL} model}} &
\multicolumn{3}{c|}{\textbf{DUC 2005}} & 
\multicolumn{3}{c|}{\textbf{DUC 2006}} & 
\multicolumn{3}{c|}{\textbf{DUC 2007}} \\ \cline{2-10}
\centering
 & \textbf{R-1} & \textbf{R-2} & \textbf{R-SU4} 
 & \textbf{R-1} & \textbf{R-2} & \textbf{R-SU4} 
 & \textbf{R-1} & \textbf{R-2} & \textbf{R-SU4} 
\\ \hline



\textbf{{\textit{with} BERTSUM}} & {40.4} & {9.2} & {14.7} & {43.2} & {10.7} & {16.4} & {44.6} & {12.4} & {17.7} 
\\\hline

\textbf{{\textit{with} BART}} & {40.3} & {8.7} & {14.8} & {42.9} & {10.2} & {16.0} & {44.2} & {11.4} & {17.0} 
\\\hline
\textbf{{\textit{with} PEGASUS}}  & {39.8} & {9.2} & {14.7} & \textbf{44.0} & \textbf{11.4} & {16.7} & {44.7} & {12.6} & {17.7} 
\\\hline
\textbf{{\textit{with} T5}} & \textbf{41.1} & \textbf{9.8} & \textbf{15.8} & {43.5} & {11.1} & {16.8} & {45.2} & \textbf{12.9} & {18.2} 
\\\hline
\textbf{SOTA} & 40.8 & 8.1 & 15.7 & 43.9 & 10.1 & \textbf{17.8} & \textbf{46.0} & {12.5} & \textbf{18.5}

\\\hline

\end{tabular}
\label{tab:csmdqfasf} 

\end{table*}}

\newcommand{\csmdqfasfwH}{
\begin{table*}[!h]

\renewcommand\thetable{B}
\caption{\change{Case study results in terms of F1 and Recall on the MD-QFAS datasets based on utilizing various models for weak supervision. Here, we denote `ROUGE' as 'R'.}} 
\centering

\begin{tabular}
{|c|c|c|c|c|c|c|c|c|c|}

\hline
\multicolumn{10}{|c|}{\textbf{Metric: F1 Score}}  \\ \hline
& \multicolumn{9}{|c|}{\textbf{Datasets}} \\ \cline{2-10}

\multicolumn{1}{|c|}{\textbf{PreQFAS\textsubscript{WSL} model}} &
\multicolumn{3}{c|}{\textbf{DUC 2005}} & 
\multicolumn{3}{c|}{\textbf{DUC 2006}} & 
\multicolumn{3}{c|}{\textbf{DUC 2007}} \\ \cline{2-10}
\centering
 & \textbf{R-1} & \textbf{R-2} & \textbf{R-SU4} 
 & \textbf{R-1} & \textbf{R-2} & \textbf{R-SU4} 
 & \textbf{R-1} & \textbf{R-2} & \textbf{R-SU4} 
\\ \hline

\textbf{{{with RoBERTa}}}  & \textbf{40.3} & \textbf{9.2} & \textbf{14.7} & \textbf{43.5} & 10.8 & \textbf{16.5} & \textbf{44.7} & \textbf{12.4} & \textbf{17.7}
\\ \hline

\textbf{{{with BERTSUM\textsubscript{EXT}}}} & 40.1 & 8.9 & 14.5 & 43.4 & \textbf{10.9} & \textbf{16.5} & 44.3 & 11.9 & 17.2
\\\hline

\textbf{{{with BERTSUM\textsubscript{ABS-EXT}}}} & 40.0 & 8.7 & 14.5 & 42.5 & 10.6 & 16.0 & 44.2 & 11.8 & 17.1
\\\hline

\multicolumn{10}{|c|}{\textbf{Metric: Recall}}  \\ \hline
& \multicolumn{9}{|c|}{\textbf{Datasets}} \\ \cline{2-10}

\multicolumn{1}{|c|}{\textbf{PreQFAS\textsubscript{WSL} model}} &
\multicolumn{3}{c|}{\textbf{DUC 2005}} & 
\multicolumn{3}{c|}{\textbf{DUC 2006}} & 
\multicolumn{3}{c|}{\textbf{DUC 2007}} \\ \cline{2-10}
\centering
 & \textbf{R-1} & \textbf{R-2} & \textbf{R-SU4} 
 & \textbf{R-1} & \textbf{R-2} & \textbf{R-SU4} 
 & \textbf{R-1} & \textbf{R-2} & \textbf{R-SU4} 
\\ \hline

\textbf{{{with RoBERTa}}} & \textbf{40.4} & \textbf{9.2} & \textbf{14.7} & \textbf{43.2} & 10.7 & \textbf{16.4} & \textbf{44.6} & \textbf{12.4} & \textbf{17.7}
\\ \hline

\textbf{{{with BERTSUM\textsubscript{EXT}}}} & 40.2 & 8.9 & 14.5 & 43.1 & \textbf{10.9} & 16.3 & 44.1 & 11.8 & 17.2
\\ \hline

\textbf{{{with BERTSUM\textsubscript{ABS-EXT}}}}  & 40.1 & 8.7 & 14.5 & 42.0 & 10.4 & 15.8 & 43.7 & 11.5 & 16.9 

\\\hline
\end{tabular}

\label{tab:csmdqfaswH} 

\end{table*}}

\newcommand{\csmdqfasansI}{
\begin{table*}[!h]
\renewcommand\thetable{C}
\caption{\change{Case study results in terms of F1 and Recall on the MD-QFAS datasets based on using different datasets to fine-tune (FT) the answer selection model. Here, we denote `ROUGE' as 'R'.}} 

\centering

\begin{tabular}
{|c|c|c|c|c|c|c|c|c|c|}
\hline
\multicolumn{10}{|c|}{\textbf{Metric: F1 Score}}  \\ \hline
& \multicolumn{9}{|c|}{\textbf{Datasets}} \\ \cline{2-10}

\multicolumn{1}{|c|}{\textbf{PreQFAS\textsubscript{WSL} model}} &
\multicolumn{3}{c|}{\textbf{DUC 2005}} & 
\multicolumn{3}{c|}{\textbf{DUC 2006}} & 
\multicolumn{3}{c|}{\textbf{DUC 2007}} \\ \cline{2-10}
\centering
 & \textbf{R-1} & \textbf{R-2} & \textbf{R-SU4} 
 & \textbf{R-1} & \textbf{R-2} & \textbf{R-SU4} 
 & \textbf{R-1} & \textbf{R-2} & \textbf{R-SU4} 
\\ \hline

\textbf{{FT on MS-MARCO}} & \textbf{40.3} & \textbf{9.2} & \textbf{14.7} & \textbf{43.5} & \textbf{10.8} & \textbf{16.5} & \textbf{44.7} & \textbf{12.4} & \textbf{17.7}
\\\hline

\textbf{{FT on TREC-QA}} & 39.9 & {8.9} & 14.5 & 42.3 & {10.3} & 15.8 & 43.9 & {11.5} & 16.9
\\\hline

\textbf{{{FT on Wiki-QA}}} & 39.6 & {8.7} & 14.2 & 42.5 & {10.3} & 15.9 & 43.4 & {11.4} & 16.7
\\\hline
\textbf{{{FT on SemEval (2015)}}} & 39.6 & {8.6} & 14.2 & 42.8 & {10.2} & 15.9 & 43.9 & {11.5} & 16.7
\\\hline
\textbf{{{FT on SemEval (2016-17)}}} & 40.0 & {8.7} & 14.4 & 42.9 & {10.3} & 15.9 & 44.4 & {11.8} & 17.0

\\\hline

\multicolumn{10}{|c|}{\textbf{Metric: Recall}}  \\ \hline
& \multicolumn{9}{|c|}{\textbf{Datasets}} \\ \cline{2-10}

\multicolumn{1}{|c|}{\textbf{PreQFAS\textsubscript{WSL} model}} &
\multicolumn{3}{c|}{\textbf{DUC 2005}} & 
\multicolumn{3}{c|}{\textbf{DUC 2006}} & 
\multicolumn{3}{c|}{\textbf{DUC 2007}} \\ \cline{2-10}
\centering
 & \textbf{R-1} & \textbf{R-2} & \textbf{R-SU4} 
 & \textbf{R-1} & \textbf{R-2} & \textbf{R-SU4} 
 & \textbf{R-1} & \textbf{R-2} & \textbf{R-SU4} 
\\ \hline

\textbf{{FT on MS-MARCO}} & \textbf{40.4} & \textbf{9.2} & \textbf{14.7} & \textbf{43.2} & \textbf{10.7} & \textbf{16.4} & \textbf{44.6} & \textbf{12.4} & \textbf{17.7}
\\\hline

\textbf{{FT on TREC-QA}} & 39.9 & {8.9} & 14.5 & 42.1 & {10.2} & 15.7 & 43.8 & {11.5} & 16.8
\\\hline

\textbf{{{FT on Wiki-QA}}} & 39.6 & {8.7} & 14.2 & 42.3 & {10.3} & 15.8 & 43.3 & {11.4} & 16.6
\\\hline
\textbf{{{FT on SemEval (2015)}}} & 39.6 & {8.6} & 14.2 & 42.6 & {10.1} & 15.8 & 43.7 & {11.4} & 16.6
\\\hline
\textbf{{{FT on SemEval (2016-17)}}} & 40.0 & {8.8} & 14.4 & 42.9 & {10.2} & 15.8 & 44.3 & {11.7} & 17.0

\\\hline
\end{tabular}

\label{tab:csmdqfasansI} 

\end{table*}}

\newcommand{\csmdqfash}{
\begin{table*}

\caption{\change{Human evaluation results in terms of Coherence (C), Fluency (F), and Informativeness (I). }}
\begin{tabular}
{|c|c|c|c|c|c|c|c|c|c|}

\hline
\centering

& \multicolumn{9}{|c|}{\textbf{Datasets}} \\ \cline{2-10}

\multicolumn{1}{|c|}{\textbf{Models}} &
\multicolumn{3}{c|}{\textbf{DUC 2005}} & 
\multicolumn{3}{c|}{\textbf{DUC 2006}} & 
\multicolumn{3}{c|}{\textbf{DUC 2007}} \\ \cline{2-10}
\centering
 & \textbf{C} & \textbf{F} & \textbf{I} 
 & \textbf{C} & \textbf{F} & \textbf{I} 
 & \textbf{C} & \textbf{F} & \textbf{I} 
\\ \hline

\textbf{PreQFAS\textsubscript{SFT} - BERTSUM} 
& {3.42} & {3.34} & {3.61} 
& {3.70} & {3.73} & {4.07} 
& {3.37} & {3.33} & {3.70} 

\\\hline

\textbf{PreQFAS\textsubscript{WSL} - BERTSUM}
& {3.63} & {3.73} & {3.63} 
& {3.93} & {3.70} & {3.97} 
& {3.87} & {3.77} & {3.83} 
\\\hline

\textbf{PreQFAS\textsubscript{WSL} - BART}
& \textbf{4.23} & \textbf{4.20} & \textbf{4.49} 
& \textbf{4.50} & \textbf{4.43} & \textbf{4.57} 
& {4.01} & \textbf{4.17} & {4.37} 
\\\hline
\textbf{{PreQFAS\textsubscript{WSL} - PEGASUS}}
& {3.87} & {4.13} & {4.11} 
& {4.23} & {4.17} & {4.40} 
& \textbf{4.23} & {4.07} & {4.43} 
\\\hline
\textbf{{PreQFAS\textsubscript{WSL} - T5}}
& {3.90} & {4.17} & {4.31} 
& {4.11} & {4.20} & {4.23} 
& {4.13} & \textbf{4.17} & \textbf{4.53} 

\\\hline

\end{tabular}
\label{tab:csmdqfashuman} 
\end{table*}}

\newcommand{\dbpediaanalysis}{
\begin{table*}

\caption{Debatepedia dataset analysis based on a randomly sampled 100 examples.}

\centering

\begin{tabular}{|l|c|}
\hline

\textbf{Analysis Type} & \textbf{Result}
\\ \hline

\textbf{Queries having no relevance with the documents or the summaries} & 52\%
\\ \hline

\textbf{Queries are Yes/No type close-ended questions} & 70\%
\\ \hline

\textbf{Queries are relevant to the documents but the summaries are more of generic} & 81\%
\\ \hline

\end{tabular}

\label{tab:dbpediaanalysis} 

\end{table*}}

\newcommand{\zssdqfasGo}{
\begin{table*}[!h]
\renewcommand\thetable{A}
\caption{Case study results on the Debatepedia dataset while using different datasets for pre-training. Here, `R', `P', and `F' denote `Recall', `Precision', and `F1' respectively.} 
\centering

\begin{tabular}
{|c|c|c|c|c|c|c|c|c|c|}
\hline
& \multicolumn{9}{|c|}{\textbf{Metrics}} \\ \cline{2-10}

\multicolumn{1}{|c|}{\textbf{MODEL}} &
\multicolumn{3}{c|}{\textbf{ROUGE-1}} & 
\multicolumn{3}{c|}{\textbf{ROUGE-2}} & 
\multicolumn{3}{c|}{\textbf{ROUGE-L}} \\ \cline{2-10}

 & \textbf{R} & \textbf{P} & \textbf{F} 
& \textbf{R} & \textbf{P} & \textbf{F} 
& \textbf{R} & \textbf{P} & \textbf{F} 
\\ \hline

\textbf{PreQFAS\textsubscript{XSUM\textsubscript{}}} & 58.0 & \textbf{60.4} & 58.5 & 45.2 & \textbf{46.1} & 45.5 & 57.1 & \textbf{59.3} & 57.7
\\\hline

\textbf{PreQFAS\textsubscript{CNN-DM\textsubscript{}}} & \textbf{59.0} & 60.3 & \textbf{59.3} & \textbf{45.4} & 46.0 & \textbf{45.6} & \textbf{57.9} & 59.2 & \textbf{58.2}
\\\hline

\end{tabular}
\label{tab:zsqfasGo} 

\end{table*}}

\newcommand{\zssdqfasG}{
\begin{table*}[!h]
\renewcommand\thetable{A}
\caption{Case study results based on the F1 metric on MediQA-AnS and Debatepedia while using different datasets for pre-training. Here, we denote `ROUGE' as 'R'.} 

\centering

\begin{tabular}
{|c|c|c|c|c|c|c|c|c|c|}
\hline
& \multicolumn{9}{|c|}{\textbf{Datasets}} \\ \cline{2-10}

\multicolumn{1}{|c|}{\textbf{MODEL}} &
\multicolumn{3}{c|}{\textbf{MediQA-AnS (Pages)}} & 
\multicolumn{3}{c|}{\textbf{MediQA-AnS (Passages)}} & 
\multicolumn{3}{c|}{\textbf{Debatepedia}} \\ \cline{2-10}
\centering
 & \textbf{R-1} & \textbf{R-2} & \textbf{R-L} 
& \textbf{R-1} & \textbf{R-2} & \textbf{R-L} 
& \textbf{R-1} & \textbf{R-2} & \textbf{R-L} 
\\ \hline

\textbf{PreQFAS\textsubscript{XSUM}} & {23.07} & {5.41} & {15.35} & {29.89} & {11.29} & {21.05} & {58.5} & {45.5} & {57.7}

\\\hline
\textbf{PreQFAS\textsubscript{CNN-DM}} & \textbf{25.30} & \textbf{7.47} & \textbf{17.53} & \textbf{33.19} & \textbf{15.49} & \textbf{24.80} & \textbf{59.3} & \textbf{45.6} & \textbf{58.2}
\\ \hline

\end{tabular}
\label{tab:zsqfasG}

\end{table*}}

%% file: figures.tex
\newcommand{\canaifigure}{
\begin{figure*}[t!]
\begin{center}
\includegraphics[width=\textwidth]{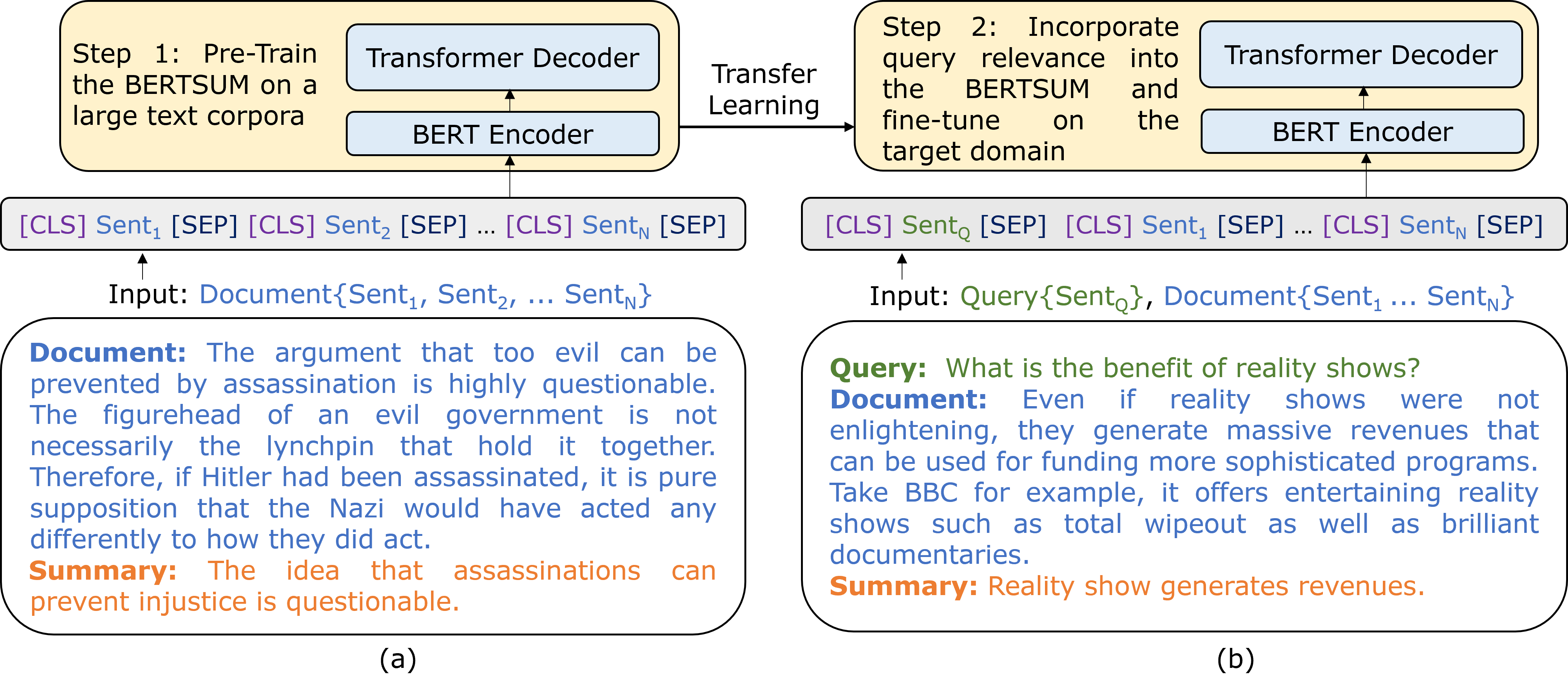}

\caption[Proposed Model: Single-Document Query Focused Abstractive Summarization task]{Our proposed PreQFAS model works in two steps: (a) Pre-train the BERTSUM model on a generic abstractive summarization corpus and (b) Fine-tune the pre-trained model for the QFAS task on the target domain.}
\label{fig1}
\end{center}
\end{figure*}}

\newcommand{\attention}{
\begin{figure*}[t!]
\begin{center}
\includegraphics[width=\linewidth]{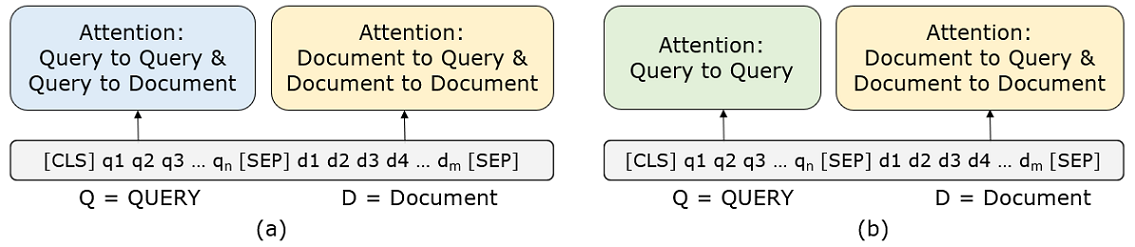}
\caption[An overview of various attention models]{
An overview of various attention models. (a) The Bidirectional Self-Attention Mechanism. (b) The Query-Document Attention Mechanism. 
}
\label{fig:AttentionModelOverview}
\end{center}
\end{figure*}}

\newcommand{\mdqfasfilterfigure}{
\begin{figure*}[t!]
\begin{center}
\includegraphics[width=\textwidth]{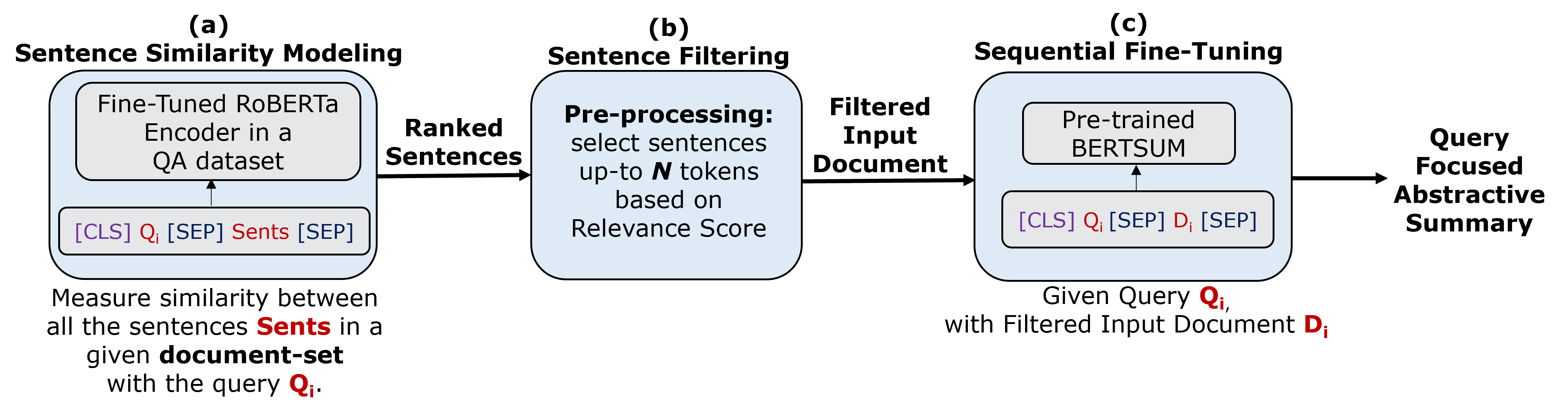}

\caption{The proposed PreQFAS model for long sequences (i.e., multi-document scenarios) based on Sentence Filtering and Sequential Fine-Tuning: (a) First, all sentences in a document set are ranked by measuring their similarity score with the query. (b) Then, these ranked sentences are combined together to create a filtered input (up-to $n$ tokens). (c) Finally, the query relevance is incorporated into the filtered input document and then given as input to the pre-trained BERTSUM model for sequential fine-tuning.}
\label{filterfig}
\end{center}
\end{figure*}}

\newcommand{\mdqfasfilterfigureexample}{
\begin{figure*}[t!]
\begin{center}
\includegraphics[width=\textwidth]{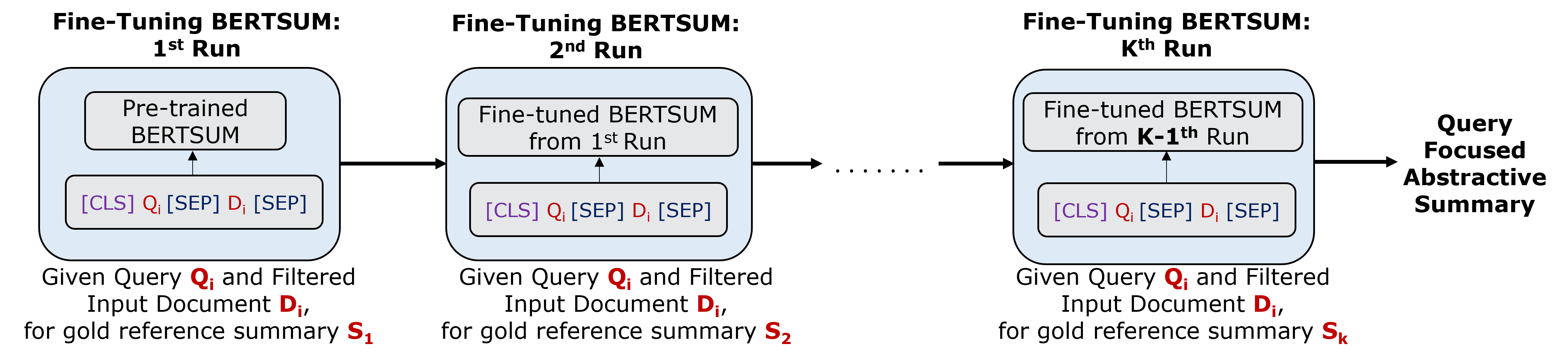}

\caption{Sequential fine-tuning of the BERTSUM model using $K$ different gold summaries. For $K$ gold summaries, the model will be fine-tuned $K$ times (i.e., $K$ fine-tuning runs).}
\label{filterexamplefig}
\end{center}
\end{figure*}}

\newcommand{\distantfigureroberta}{
\begin{figure*}[t!]
\begin{center}
\includegraphics[width=\linewidth]{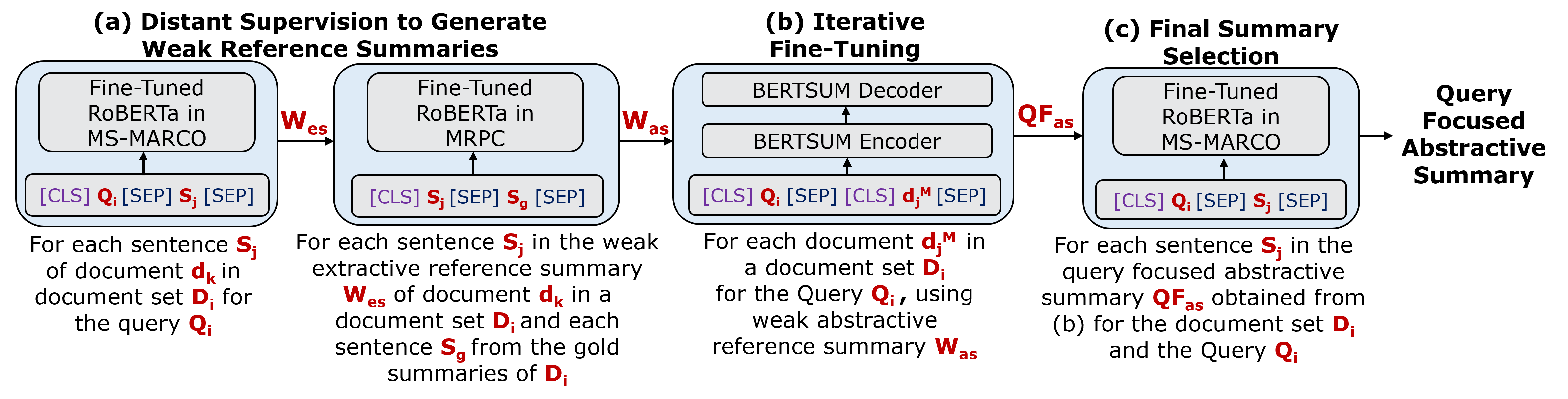}

\caption[Proposed Model: Multi-Document Query Focused Abstractive Summarization task]{
An overview of our proposed PreQFAS model for long text sequences (i.e., multi-document scenarios) that uses the fine-tuned \textbf{RoBERTa\textsubscript{MS-MARCO}} model to (a) generate the initial weak extractive reference summary of each document followed by utilizing the \textbf{RoBERTa\textsubscript{MRPC}} model for distant supervision to generate the weak abstractive reference summary. Then, (b) the pre-trained \textbf{BERTSUM} model is fine-tuned to iteratively generate the query focused abstractive summary of each document. Finally, all the generated query focused abstractive summaries are (c) ranked by the \textbf{RoBERTa\textsubscript{MS-MARCO}} model to select the final summary. 
}

\label{fig:ModelOverview}
\end{center}

\end{figure*}}

\newcommand{\seqvsbase}{
\begin{figure*}[t!]
\begin{center}
\includegraphics[width=12cm,height=14cm]{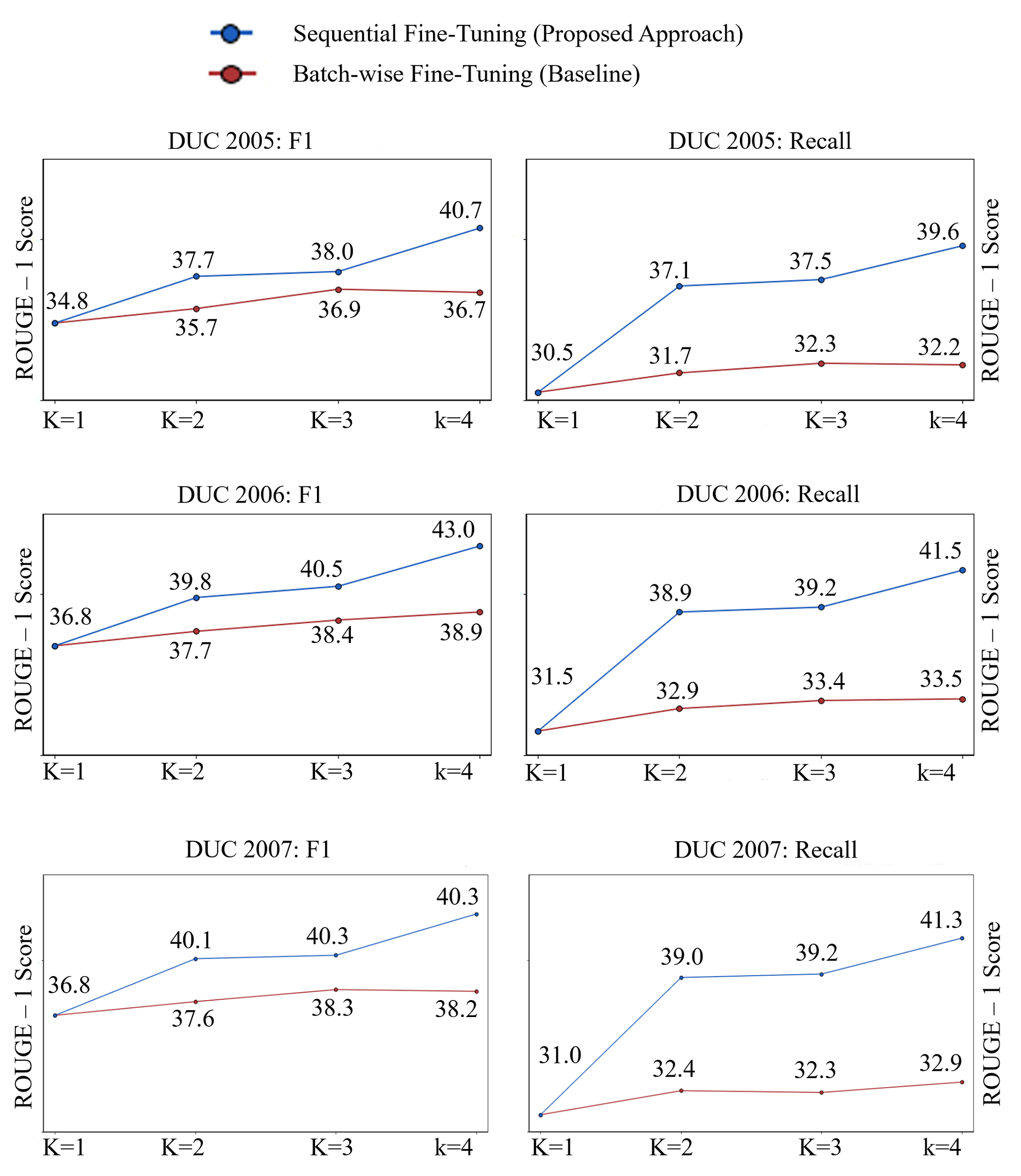}

\caption{Case study results in terms of ROUGE-1 on various MD-QFAS datasets by varying the total number of gold reference summaries $K$ used for sequential fine-tuning of the proposed PreQFAS\textsubscript{SFT} model and batch-wise fine-tuning of the baseline BERTSUM model.}

\label{fig:SeqVsBase}
\end{center}

\end{figure*}}